\documentclass[11pt]{article}

\usepackage{graphicx}
\usepackage{amssymb}
\usepackage{physics}
\usepackage{amsthm}
\usepackage{multirow}
\usepackage{comment}

\usepackage{rotating}

\usepackage{microtype} % microtypography
\usepackage{booktabs}  % tables
\usepackage{url}  % urls

\usepackage{amsmath}
\usepackage{amsthm}

\usepackage{pifont}% http://ctan.org/pkg/pifont

\newcommand{\keypoint}[1]{\noindent\textbf{#1:}}
\newcommand{\rebuttal}[1]{\textcolor{black}{#1}}
\newcommand{\doublecheck}[1]{#1}
\newcommand{\cut}[1]{}

\newcommand{\ques}[1]{\vspace{0.1cm}\noindent\textbf{#1}\quad}

\newcommand{\tb}[1]{\textbf{#1}}

\usepackage{algorithm}
\usepackage{algpseudocode}

\usepackage{amsmath,amsfonts,bm}

\def\eqref#1{equation~\ref{#1}}
\def\1{\bm{1}}

\def\vone{{\bm{1}}}

\def\vtheta{{\bm{\theta}}}
\def\vphi{{\bm{\phi}}}

\def\vh{{\bm{h}}}

\def\vp{{\bm{p}}}

\def\vs{{\bm{s}}}
\def\vt{{\bm{t}}}

\def\vx{{\bm{x}}}
\def\vy{{\bm{y}}}

\def\evp{{p}}

\def\mP{{\bm{P}}}

\def\mS{{\bm{S}}}
\def\mT{{\bm{T}}}

\def\mX{{\bm{X}}}

\DeclareMathAlphabet{\mathsfit}{\encodingdefault}{\sfdefault}{m}{sl}
\SetMathAlphabet{\mathsfit}{bold}{\encodingdefault}{\sfdefault}{bx}{n}

\def\gA{{\mathcal{A}}}

\def\gD{{\mathcal{D}}}

\def\gX{{\mathcal{X}}}
\def\gY{{\mathcal{Y}}}

\def\sH{{\mathbb{H}}}

\DeclareMathOperator*{\argmax}{arg\,max}
\DeclareMathOperator*{\argmin}{arg\,min}

\usepackage{times}
\usepackage{epsfig}

\usepackage[final]{automl}
\usepackage[numbers]{natbib}
\setcitestyle{square}

\usepackage[capitalize]{cleveref}
\crefname{section}{Sec.}{Secs.}
\Crefname{section}{Section}{Sections}
\Crefname{table}{Table}{Tables}
\crefname{table}{Tab.}{Tabs.}

\title{Better Practices for Domain Adaptation}

\author[1]{\nameemail{Linus Ericsson}{linus.ericsson@ed.ac.uk}}
\author[2]{\nameemail{Da Li}{da.li1@samsung.com}}
\author[1,2]{\nameemail{Timothy M. Hospedales}{t.m.hospedales@ed.ac.uk}}

\affil[1]{University of Edinburgh}
\affil[2]{Samsung AI Center Cambridge}

\hypersetup{%
  pdfauthor={}, % will be reset to "Anonymous" unless the "final" package option is given
  pdftitle={},
  pdfsubject={},
  pdfkeywords={}
}

\begin{document}

\maketitle

%%%%%%%%% ABSTRACT
\begin{abstract}
Distribution shifts are all too common in real-world applications of machine learning. Domain adaptation (DA) aims to address this by providing various frameworks for adapting models to the deployment data without using labels. However, the domain shift scenario raises a second more subtle challenge: the difficulty of performing hyperparameter optimisation (HPO) for these adaptation algorithms without access to a labelled validation set. % as typically assumed in supervised learning.
The unclear validation protocol for DA has led to bad practices in the literature, such as performing HPO using the target test labels when, in real-world scenarios, they are not available. This has resulted in over-optimism about DA research progress compared to reality.
In this paper, we analyse the state of DA when using good evaluation practice, by benchmarking a suite of candidate validation criteria and using them to assess popular adaptation algorithms. We show that there are challenges across all three branches of  domain adaptation methodology including Unsupervised Domain Adaptation (UDA), Source-Free Domain Adaptation (SFDA), and Test Time Adaptation (TTA). 
%In each case, there is a large gap between oracle HPO and achievable performance given real computable validators. 
While the results show that realistically achievable performance is often worse than expected, they also show that using proper validation splits is beneficial, as well as showing that some previously unexplored validation metrics provide the best options to date. 
%\doublecheck{Additionally we highlight the importance of using proper validation splits in order to reliably estimate target generalisation performance. Finally, we find previously unexplored validation metrics that are widely applicable across all settings.} 
Altogether, our improved practices covering data, training, validation and hyperparameter optimisation form a new rigorous pipeline to improve benchmarking, and hence research progress, within this important field going forward.
\end{abstract}

%%%%%%%%% BODY TEXT
\section{Introduction}
\label{sec:intro}
Supervised deep learning models achieve impressive results when training and testing data are identically distributed. However, perhaps the main failure mode of computer vision and pattern recognition systems in practice is due to the near-ubiquitous distribution shift between data curated for model training, and real-world data encountered during deployment \cite{csurka2022visual}. This distribution shift issue has motivated a tremendous amount of work in the area of unsupervised domain adaptation (UDA) \cite{csurka2022visual}. UDA methods aim to alleviate domain shift by collecting freely available unlabelled data during deployment to a target domain and adapting vision models based on this unlabelled data. 

Hundreds of unsupervised adaptation algorithms have now been proposed based on various principles from distribution alignment \cite{Long2015LearningNetworks}, to domain adversarial learning \cite{ganin2016domainAdversarialNN} and much more. However, without exception, a key challenge for every one of these algorithms is: \emph{how do we tune hyperparameters and conduct model selection?} In conventional supervised learning, hyperparameters and model selection (stopping criteria) are handled systematically by maximising accuracy on a validation split of the training set. In unsupervised domain adaptation there is no such straightforward solution because the target domain has no labels with which to compute accuracy, and the source domain is not representative of the target domain. 

Despite the importance of this issue---upon which any practical application of domain adaptation hinges---there has been relatively little systematic study of validation protocols and algorithms for UDA \cite{you2019modelSelDA,cui2020batchNuclearNorm,saito2021tune}. Worse, a recent meta-review and re-evaluation of the domain adaptation literature found that most published code did not use consistent or fair model selection criteria \cite{Musgrave2020UnsupervisedCheck}, and furthermore when evaluated under consistent and fair model selection criteria most existing results can not be replicated \cite{Musgrave2020UnsupervisedCheck}. This mini ``replication crisis'' in domain adaptation highlights the need for studying validation protocols for UDA, and for fair benchmarking to drive reliable progress.

The few existing fair model selection criteria for UDA are based on diverse intuitions such as simply applying UDA algorithm objectives on the validation split of the unlabelled target set, priors on the expected distribution of labels \cite{cui2020batchNuclearNorm,saito2021tune}, or relying on the validation accuracy in the source domain \cite{you2019modelSelDA}. However there is little first principles justification to pick among these reasonable intuitions, and there is little empirical evaluation to understand which are best, and how close they come to the performance of an oracle validator, which has been the basis of many reported results in the literature \cite{Musgrave2020UnsupervisedCheck}.

%; it is unclear what assumptions they rely on and whether they hold in practice, and none of these methods were found to be reliably representative of test performance in practice \cite{Musgrave2020UnsupervisedCheck}.

%A further limitation of the existing validation criteria for UDA is that they are mostly only suited to validating adaptation of classification models. The wider world of vision problems including regression \cite{lathuiliere2019deepRegressionSurvey} and beyond are not supported. 

%Our first contribution in this paper is to develop \emph{principled} validation criteria based on bounds on the target domain risk. Importantly, our validation criteria are general bounds that can be instantiated for different kinds of learning problems. We demonstrate an instantiation of our criterion for both classification and regression problems. 

These challenges exist throughout the domain adaptation literature. They arise across all three popular branches of adaptation for recognition: Unsupervised Domain Adaptation (UDA) \cite{ganin2016domainAdversarialNN, coral, mmd}, Source-Free Domain Adaptation (SFDA) \cite{liang2020sourceFree, yang2022attracting} and Test Time Adaptation (TTA) \cite{wang2021tent, ttt_plus_plus}. 
They also arise across different kinds of domain adaptive learning problems from classification \cite{ganin2016domainAdversarialNN} to regression \cite{da_regression}, dense prediction \cite{da_segmentation}, and detection \cite{da_detection}.

\rebuttal{The lack of a clear validation criterion for DA is an obstacle to its practical application. As an example, AutoML is a field with great potential to automate machine learning tasks for real-world applications \cite{He2019AutoML:State-of-the-Art}. But in order to automate anything (e.g.~algorithm, hyperparameter, checkpoint selection), we need a metric to optimise. In the case of supervised learning, this metric is naturally validation performance on an unseen labelled set. However, the choice of metric is not straightforward for UDA/SFDA/TTA due to the lack of labels. A major contribution of this paper is to clarify what such an optimisation metric should be for domain adaptation, thereby laying the foundations that allow bringing AutoML to DA.
%Towards this, we have identified some problem settings and metrics where the validators work sufficiently well such that AutoML might work already today (UDA); and some others where the validation metrics are insufficient, meaning AutoML can’t work until better validators are developed (TTA on big domain shift).
}

To address this issue, we conduct a large-scale benchmark of 10 domain adaptation algorithms with 15 different validation criteria and three DA settings (UDA, SFDA, TTA). We identify which DA validators can be applied to each setting, characterise the size of the challenge in each case in terms of the gap between practically achievable and best-case DA performance, and identify the best existing validators. We identify effective practices in terms of using validation splits to estimate target performance. We highlight the risk of adaptation failure in SFDA and TTA as a likely fatal blocker for deployment in practice as existing validators do not reliably prevent this. These results should drive future practice both in DA research -- which should use these validators, rather than unrealistic oracle HPO; and in validator research -- which should aim to develop validators which surpass the best that we report.

%We show that (i) our new criteria exhibit the strongest correlation with testing performance to date, and (ii) performing model selection and hyperparameter optimisation with Tour criteria leads to state of the art performance among existing fair validation criteria. We go beyond \cite{Musgrave2020UnsupervisedCheck} to demonstrate the impact of validation criteria on domain adaptive regression, and disentangle the role of validation criteria on model selection vs algorithm selection. 

\section{Related Work}
\label{sec:related}
%Most similar to our work is \cite{Nath2019UM-Adapt:Distillation} whose setting includes a source domain with labels for multiple tasks (depth, segmentation and normals) and an unlabelled target. Their goal is to perform well in all three tasks on the target domain. \cite{ZamaRamirez2019LearningDomains} tackle a similar setting, but where the target only has labels for one task though we wish to solve all.
\subsection{Domain Adaptation} There are now too many domain adaptation algorithms to review here, and we refer the reader to good surveys such as \cite{csurka2022visual,patel2015vdaSurvey}. Most deep UDA algorithms proceed by performing supervised learning on the source domain data, and some kind of unsupervised objective on the target domain data. Representative families of approach include objectives that penalise misalignment between the source and target domain feature distributions \cite{Long2015LearningNetworks}, train a domain classifier that can then be used adversarially to penalise distinguishable source and target domain features \cite{ganin2016domainAdversarialNN}, or penalise deviation from a prior on the expected target label distribution \cite{Shi2012Information-TheoreticalAdaptation}. However, all algorithms have a number of hyperparameters, such as stopping iteration and strength of the weighting factor for supervised vs unsupervised loss components. How to set these hyperparameters is not clear given the lack of a labelled target domain validation set in UDA applications.

The long-established mainstream setting for unsupervised domain adaptation (UDA) assumes that source and target data are accessed simultaneously for training. Two related problem variants have more recently gained rapid popularity, namely source-free domain adaptation (SFDA) and Test Time Adaptation (TTA). SFDA refers to the condition where pre-trained source models should be adapted to the target data without revisiting the source data  \cite{liang2020sourceFree} -- for example, by unsupervised fine-tuning. TTA \cite{wang2021tent,sun2020testTrain} similarly adapts a pre-trained model without access to the source data, but assumes that the test data arrives in mini-batches, providing the opportunity to adapt to each mini-batch before making decisions on their labels. The newer SFDA and TTA have both rapidly gained traction as being more "practical" in an era of pre-trained models \cite{bommasani2021foundation}. However, algorithms for both of these settings still have many hyperparameters (e.g., learning rate, number of iterations, regularisation strengths), and hence suffer from the lack of a clear validation protocol in a DA context. Most of the seminal studies in this area do not show valid HPO criteria in their papers or code.

\subsection{Validation Approaches for DA} Comparatively few papers have systematically studied validation criteria for UDA, given the importance of this issue for its practical application. Typical solutions applied by UDA algorithm papers include: (1) Oracle risk. Many papers use the target test set for hyperparameter selection \cite{Musgrave2020UnsupervisedCheck}, which is obviously incorrect as it can not be used in real applications; (2) Source risk. Evaluating the adapted model on the source validation set is reasonable but may not be a good validation criterion due to domain shift between source and target domains; (3) Evaluating another UDA algorithm objective (such as InfoMax \cite{Shi2012Information-TheoreticalAdaptation} and MMD \cite{Long2015LearningNetworks}) on an unlabelled validation split of the target set; (4) Validation domain. Use of a held-out labelled validation domain, as used in the VisDA challenge \cite{peng2018visda}, is fair. However, this assumes multiple labelled domains, which may not be available in practice, and also raises additional questions of whether the optimal hyperparameters for the validation domain are representative of the optimal hyperparameters for the target domain.

Besides the above strategies, a few purpose-designed validation criteria have been proposed: Deep embedded validation (DEV) \cite{you2019modelSelDA} weights the source validation risk by the probability that each sample belongs to the target domain. Meanwhile, Silhouette score \cite{robbiano2022classSS}, batch nuclear-norm minimisation (BNM) \cite{cui2020batchNuclearNorm}, and soft neighbourhood density (SND) \cite{saito2021tune} criteria boil down to evaluating the adapted models' posterior label distribution on the target domain under different notions of a prior for the expected target domain label distribution. Mean ensemble-based validation (ENS) \cite{robbiano2022classSS} considers a linear combination of the above criteria. However, overall it is unclear which to prefer for DA.

%validation criteria for UDA is still an open question: It's unclear how to choose among these criteria, and empirical evaluations have shown that none of them are very effective as model selection criteria \cite{Musgrave2020UnsupervisedCheck}. Furthermore, most of these criteria are only suited to classification tasks and do not directly extend to the broader space of vision problems including regression \cite{lathuiliere2019deepRegressionSurvey}, segmentation \cite{csurka2021ssegSurvey}, etc.

\subsection{Benchmarking Domain Adaptation} There have been two major benchmarking exercises in UDA. The VisDA competition challenge \cite{peng2018visda} provides a labelled validation domain for model selection and hyperparameter optimisation (HPO). However, validation domains may not be available in practice -- and if they are, they may not be representative of the target domain. Thus, the vast majority of research literature on UDA has not used this approach. 
A recent empirical evaluation  \cite{Musgrave2020UnsupervisedCheck,musgrave2022benchmarkingUDA} analysed the GitHub repositories of a number of UDA methods and found that: (1) In practice different methods used very different validation criteria for empirical evaluation, making published results incomparable with each other; (2) A large number of prior studies used the oracle risk as a validation criterion, meaning that their results are not representative of how well domain adaptation would work in reality using validation criteria that can be implemented in practice; (3) Variation in existing validation criteria was high compared to variation across adaptation algorithms, and none of them was strongly correlated with recognition performance. Our evaluation extends this early study but goes beyond it in considering all three major branches of DA research (UDA, SFDA, TTA), exploring a wider variety of validators, and demonstrating how validator performance can be improved through proper use of validation splits within the target domain.

\section{Background}
% Introduce setting and notation.
\subsection{Problem Setup}
\keypoint{Unsupervised Domain Adaptation} In the UDA setup, one typically trains a model $f_\vtheta: \gX \mapsto \gY$ on a labelled dataset, $\gD_S = \{\vx_i, \vy_i\}_{i=1}^{N_S}$, consisting of data sampled from a source domain, $p_S$. The goal is then to adapt $f_\vtheta$ using an unlabelled dataset, $\gD_T = \{\vx_i\}_{i=1}^{N_T}$, sampled from a target domain, $p_T$. The general learning objective to be minimised w.r.t. to $\vtheta$ can be simplified as follows,
\begin{equation}
  L(f_\vtheta, \gD_S, \gD_T) = L_{\text{sup}}(f_\vtheta, \gD_S) + L_{\text{da}}(f_\vtheta, \gD_S, \gD_T),
\end{equation}
where $L_{\text{sup}}(\cdot)$ could be cross-entropy loss for classification and mean square error for regression problems, and $L_{\text{da}}(\cdot)$ is the adaptation loss, such as MMD~\cite{mmd}, CORAL~\cite{coral} and DANN~\cite{ganin2016domainAdversarialNN} losses.

\keypoint{Source-Free Domain Adaptation} The SFDA setting aims to adapt a pre-trained source domain model to the target domain, relaxing the assumption of joint occurrence of source and target domain data in UDA. So first, a source model will be optimized using source domain data: $\hat{\theta} = \underset{\theta}{\argmin}~L_{\text{sup}}(f_\vtheta, \gD_S)$. Then the trained source model $\hat{\theta}$ will be adapted to the target domain by
\begin{equation}
  {\theta}^* = \underset{\hat{\theta}}{\argmin}~L_{\text{sfda}}(f_{\hat{\theta}}, \gD_T).
\end{equation}
where now $L_{\text{sfda}}$ is an unsupervised loss, such as clustering~\cite{yang2022attracting} or information maximization~\cite{liang2020sourceFree}.

\keypoint{Test-Time Adaptation} Unlike SFDA, TTA assumes the batch-wise target domain data $X\sim\gD_T$ comes in a stream and adapts a pre-trained source model for each minibatch $X$ as 
\begin{equation}
  {\theta}^* = \underset{\hat{\theta}}{\argmin}~\underset{X\sim\gD_T}{L_{\text{tta}}}(f_{\hat{\theta}}, X),
\end{equation}
where $L_{\text{tta}}$ is commonly the unsupervised loss, such as self-supervised learning and entropy minimisation losses, which could be essentially similar to $L_{\text{sfda}}$.

\begin{figure}[t]
\centering
\includegraphics[width=1.0\textwidth]{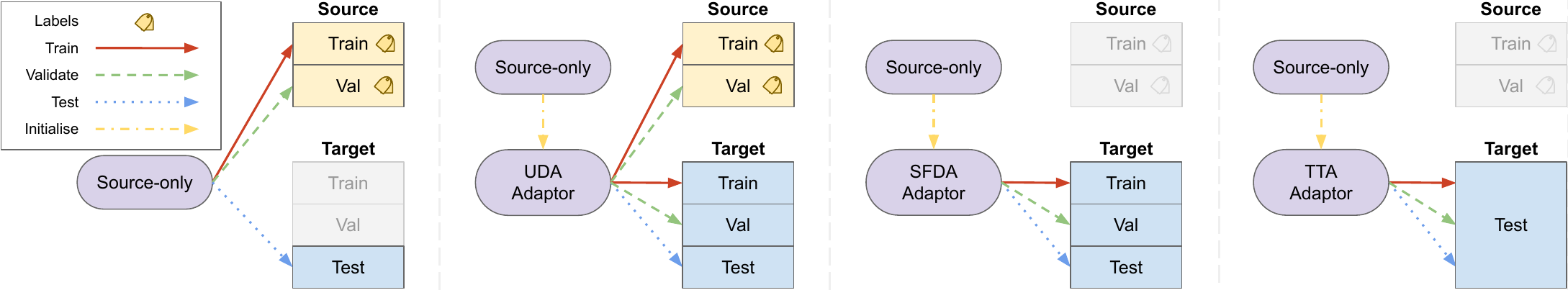}
\caption{How the source and target domains are split and how each split is used for (1) the source-only model (2) UDA adaptors, (3) SFDA adaptors and (4) TTA adaptors.}
\label{fig:domain_split_and_use_diagram}
\end{figure}

\subsection{Model Selection}
Due to the lack of target domain labels in the various domain adaptation settings we consider, the model selection process must proceed as follows. Given a set of candidate models, as configured by hyperparameters $\vh \in \sH$, where $\sH$ is the pool of hyperparameter sets, the best candidate model is selected based on its evaluation score, $d(f_\vtheta, \gD_V)$\footnote{Assuming the model performance is a monotonically decreasing function of the output of $d(\cdot, \gD_V)$.}, where $\gD_V$ is a validation dataset. The process can be formalised as
\begin{equation}
\begin{aligned}
 \vh^* &= \argmax_{\vh}~d(f_{\vtheta^*_\vh}, \gD_V), \\
 \textup{s.t.} \quad \vtheta^*_\vh &= \argmin_{\vtheta}~L(f_\vtheta, \gD_S, \gD_T; \vh).
\end{aligned}
\end{equation}
However, two things in UDA complicate this process: 1) choice of the validation set $\gD_V$; and 2) definition of the evaluation metric $d(\cdot,\cdot)$ when $\gD_V = \{\vx_i\}_{i=1}^{N_V}$ is an unlabelled set.

Several validators have been proposed in the literature, such as SND \cite{saito2021tune}, BNM \cite{cui2020batchNuclearNorm} and DEV \cite{you2019modelSelDA}. Additionally, it is worth remarking that popular DA losses such as IM \cite{liang2020sourceFree}, and Entropy~\cite{vu2018advent, morerio2018minimalentropy}, can also be used as validators.
%Several such validators have been proposed in the literature, from entropy to information maximisation \cite{Shi2012Information-TheoreticalAdaptation}, mutual information \cite{musgrave2022benchmarkingUDA} and soft neighbourhood density \cite{saito2021tune}.
We explore a large number of validators in addition to these, including those based on domain alignment, like MMD \cite{mmd} and CORAL \cite{coral}, clustering \cite{vmeasure} and feature matrix rank \cite{rankme}. The full list of validators we consider in shown \cref{tab:validators} with full details in \cref{sec:validators}.

\section{Evaluation}
\label{sec:evaluation}
Our evaluation extends the benchmark of \cite{Musgrave2020UnsupervisedCheck}. We make their setup more rigorous by splitting the target domain into train/val/test sets. Previous works often compute target performance on the same data that the algorithms adapt to or the same data that the validators use. This fails to properly measure generalisation performance as we will show later. Our splits and how we use them are detailed in \Cref{fig:domain_split_and_use_diagram}.

In order to compare different validation criteria, we train a large number of models across several datasets, algorithms and hyperparameter choices. We want the optimal validator to behave similarly to the target domain test performance of the corresponding algorithm. We measure the quality of each validator in two ways: 1) computing the Spearman rank correlation between validator scores and oracle test accuracy, 2) using the validator to select the best model for an algorithm/task pair and comparing the test performance of it against the best model as selected by the oracle.

\keypoint{Questions} Through the experimental evaluation below on three different settings, we aim to answer the following questions: (i) \emph{Are the validation criteria sufficiently good to drive HPO and model selection in UDA?}
%How well do the various validation criteria correlate with true testing performance?} 
We also extend this question to regression problems in Appendix~\ref{sec:regression}. %(ii) \emph{Which validation criterion leads to the best generalisation performance when used for model selection?}
(ii) \emph{What is the impact of validating on the training set versus an independent validation split? } (iii) \emph{Are the observations still consistent when source data is absent during adaptation (SFDA), and when we must adapt to the test-set itself (TTA)?} %(iv) And target data in streaming will have an impact or not?} %(vi) Are conclusions still valid beyond the image classification task?} 

\cut{
\begin{table}[t]
  \centering
  \caption{How the source and target domains are split and how each split is used for (1) the source-only model and (2) adaptation algorithms.}
  \label{tab:data_splits}
  \resizebox{0.6\linewidth}{!}{
  \begin{tabular}{l|llllll}
 \toprule
 {} & \multicolumn{2}{c}{Source} & & \multicolumn{3}{c}{Target} \\
 \cmidrule{2-3} \cmidrule{5-7}
 {} & Train & Val & & Train & Val & Test \\
 \midrule
 Source-only & Train & Validate & & - &- & Test \\
 Adaptors& Adapt &- & & Adapt & Validate & Test \\
 \bottomrule
  \end{tabular}
  }
\end{table}
}

\begin{table}[t]
\begin{minipage}{.5\linewidth}
  \caption{A summary of the adaptation \\ algorithms considered}
  \label{tab:algorithms}
  \centering
  \scriptsize
    \begin{tabular}{cl|l}
\toprule
& Algorithm & Approach \\
\midrule
\multirow{6}{*}{\rotatebox{90} 
    {UDA}} &
ATDOC~\cite{Liang2021DomainClassifier}& Pseudo-labelling  \\
& BNM~\cite{cui2020batchNuclearNorm} & SVD loss  \\
& DANN~\cite{ganin2016domainAdversarialNN} & Adversarial  \\
& MCC~\cite{Jin2020MinimumAdaptation} & Information maximisation \\
& MCD~\cite{Saito2018MaximumAdaptation} & Classifier discrepancy \\
& MMD~\cite{Long2015LearningNetworks} & Feature distance  \\
\midrule
\multirow{3}{*}{\rotatebox{90} 
    {SFDA}} & AAD~\cite{yang2022attracting} & Clustering \\
& NRC~\cite{yang2021exploiting} & Graph clustering\\
& SHOT~\cite{liang2020sourceFree} & Information maximisation \\
\midrule
\multirow{2}{*}{\rotatebox{90} 
    {TTA}} 
& SHOT~\cite{liang2020sourceFree} & Information maximisation \\
& TENT~\cite{wang2021tent} & Entropy minimisation \\
%& TTT++~\cite{sun2020testTrain} & Self-supervised learning \\
    \bottomrule
  \end{tabular}
\end{minipage}%
\begin{minipage}{.5\linewidth}
  \caption{A summary of the validators \\ considered.}
  \label{tab:validators}
  \centering
  \scriptsize
    \begin{tabular}{l|l}
\toprule
Criterion & Approach \\
\midrule
RankMe \cite{rankme} & Rank estimation \\
AMI \cite{musgrave2022benchmarkingUDA,robbiano2022classSS} & Cluster quality \\
ARI \cite{ari} & Cluster quality \\
V-Measure \cite{vmeasure} & Cluster quality \\
FMI \cite{fmi} & Cluster quality \\
Silhouette \cite{musgrave2022benchmarkingUDA,robbiano2022classSS} & Cluster quality \\
DBI \cite{dbi} & Cluster quality \\
CHI \cite{chi} & Cluster quality \\
BNM \cite{cui2020batchNuclearNorm}  & Label prior \\
MMD \cite{Long2015LearningNetworks} & Domain Alignment \\
CORAL \cite{coral} & Domain Alignment \\
SND \cite{saito2021tune} & Label prior \\
InfoMax \cite{Shi2012Information-TheoreticalAdaptation} & Label prior \\
Entropy & Label prior \\
Source Accuracy & Source accuracy \\
\bottomrule
  \end{tabular}
\end{minipage} 
\end{table}

\subsection{Unsupervised Domain Adaptation}
\subsubsection{Setup} In this section we describe our evaluation procedure for UDA.

\keypoint{Datasets}
We use a wide range of UDA benchmark datasets: MNIST-M \cite{ganin2016domainAdversarialNN} which consists of a domain shift from standard MNIST \cite{LeCun2010MNISTDatabase} to a modified version; The VisDA-2017 \cite{visda2017} dataset which contains \textit{train}, \textit{validation} and \textit{test} domains --- we consider the shifts \texttt{train} $\rightarrow$ \texttt{validation} and \texttt{train} $\rightarrow$ \texttt{test}; Office-31 \cite{Saenko2010AdaptingDomains} which consists of three domains: \textit{amazon}, \textit{dslr} and \textit{webcam}; and Office-Home \cite{officehome} with four domains: \textit{art}, \textit{clipart}, \textit{product} and \textit{real}. In total, we consider 21 different domain shifts.

\cut{
\begin{table*}
  \centering
  \caption{The average target test accuracy of the top 5 performing models per algorithm and task.}
  \label{tab:uda_top_5}
  \resizebox{1.0\linewidth}{!}{
  \begin{tabular}{l|c|cc|cccccc|cccccccccccc}
  \toprule
  {} &  MM &  TV &  TT &  AD &  AW &  DA &  DW &  WA &  WD &  AC &  AP &  AR &  CA &  CP &  CR &  PA &  PC &  PR &  RA &  RC &  RP \\
  \midrule
  source-only && 64.70 & 59.38 & 83.17 & 77.48 & 69.50 & 97.48 & 67.48 & 100.00 & 39.59 & 61.46 & 71.42 & 51.19 & 60.72 & 63.81 & 51.23 & 38.97 & 72.98 & 63.99 & 41.95 & 75.32 \\
  \midrule
  ATDOC  && 71.64 & 72.08 & 86.14 & 80.88 & 74.79 & 96.35 & 75.00 &  99.01 & 45.98 & 70.06 & 77.16 & 62.18 & 71.81 & 72.36 & 54.61 & 47.56 & 74.43 & 70.45 & 52.12 & 77.46 \\
  BNM&& 67.19 & 71.06 & 87.52 & 79.50 & 72.77 & 97.99 & 74.50 &  99.60 & 48.36 & 71.43 & 75.18 & 58.40 & 67.92 & 72.98 & 55.10 & 45.80 & 76.54 & 67.24 & 54.18 & 79.46 \\
  DANN&& 64.10 & 65.39 & 84.95 & 77.74 & 71.49 & 97.23 & 67.73 & 100.00 & 45.57 & 60.61 & 71.19 & 51.98 & 57.03 & 62.98 & 50.16 & 46.85 & 73.42 & 65.19 & 55.51 & 75.77 \\
  MCC&& 69.24 & 71.58 & 90.10 & 88.81 & 72.84 & 98.74 & 74.43 & 100.00 & 49.55 & 70.98 & 77.89 & 56.91 & 69.16 & 72.64 & 58.77 & 46.25 & 78.85 & 67.04 & 52.90 & 79.57 \\
  MCD&& 61.58 & 65.16 & 85.35 & 79.62 & 65.07 & 97.36 & 66.31 & 100.00 & 43.23 & 60.99 & 71.67 & 53.58 & 59.57 & 63.42 & 49.75 & 40.23 & 73.12 & 64.16 & 48.45 &  \\
  MMD&& 63.11 & 61.82 & 83.56 & 77.23 & 66.67 & 97.48 & 68.90 & 100.00 & 45.34 & 63.44 & 72.39 & 55.60 & 60.09 & 64.98 & 49.51 & 41.86 & 74.77 & 66.71 & 51.13 &  \\
  \bottomrule
  \end{tabular}
  }
\end{table*}
}

\begin{table*}[t]
\caption{Comparison of validation criteria for model selection in UDA. Averages over all 21 domain transfers evaluated. We report (i) the target test performance for the top models selected by each validator, and (ii) the correlation coefficient between the validator scores and the test performance over all hyperparameters and checkpoints. The colour of a cell indicates whether that model/validator combination beats the source-only model (green) or not (red).}
  \label{tab:uda}
\resizebox{1.0\linewidth}{!}{
  \begin{tabular}{l|ccccccccccccccc|c}
\toprule
  {}        & RankMe                        & AMI                           & ARI                           & V-Measure                     & FMI                           & Silhouette                    & DBI                           & CHI                           & BNM                           & MMD                           & CORAL                         & SND                           & IM                            & Entropy                       & Accuracy                      & Oracle
\\ \midrule
ATDOC       & \cellcolor[HTML]{FFD1D1}58.24 & \cellcolor[HTML]{E9FFC0}67.70 & \cellcolor[HTML]{E9FFC0}67.71 & \cellcolor[HTML]{E9FFC0}67.79 & \cellcolor[HTML]{E9FFC0}67.71 & \cellcolor[HTML]{FFD1D1}46.73 & \cellcolor[HTML]{FFD1D1}49.55 & \cellcolor[HTML]{FFD1D1}16.55 & \cellcolor[HTML]{E9FFC0}64.29 & \cellcolor[HTML]{FFD1D1}52.46 & \cellcolor[HTML]{FFD1D1}55.96 & \cellcolor[HTML]{FFD1D1}24.13 & \cellcolor[HTML]{E9FFC0}64.61 & \cellcolor[HTML]{FFD1D1}61.23 & \cellcolor[HTML]{E9FFC0}68.06 & \cellcolor[HTML]{E9FFC0}72.24 \\
BNM         & \cellcolor[HTML]{FFD1D1}61.36 & \cellcolor[HTML]{E9FFC0}69.32 & \cellcolor[HTML]{E9FFC0}69.48 & \cellcolor[HTML]{E9FFC0}69.29 & \cellcolor[HTML]{E9FFC0}69.48 & \cellcolor[HTML]{FFD1D1}62.42 & \cellcolor[HTML]{FFD1D1}51.58 & \cellcolor[HTML]{FFD1D1}33.10 & \cellcolor[HTML]{E9FFC0}66.88 & \cellcolor[HTML]{FFD1D1}52.64 & \cellcolor[HTML]{FFD1D1}60.92 & \cellcolor[HTML]{FFD1D1}47.70 & \cellcolor[HTML]{E9FFC0}67.01 & \cellcolor[HTML]{E9FFC0}65.98 & \cellcolor[HTML]{E9FFC0}66.02 & \cellcolor[HTML]{E9FFC0}71.09 \\
DANN        & \cellcolor[HTML]{FFD1D1}62.00 & \cellcolor[HTML]{E9FFC0}64.76 & \cellcolor[HTML]{E9FFC0}64.35 & \cellcolor[HTML]{E9FFC0}64.55 & \cellcolor[HTML]{FFD1D1}63.23 & \cellcolor[HTML]{FFD1D1}56.06 & \cellcolor[HTML]{FFD1D1}53.86 & \cellcolor[HTML]{FFD1D1}36.89 & \cellcolor[HTML]{FFD1D1}62.72 & \cellcolor[HTML]{FFD1D1}51.62 & \cellcolor[HTML]{FFD1D1}60.61 & \cellcolor[HTML]{FFD1D1}46.51 & \cellcolor[HTML]{FFD1D1}62.79 & \cellcolor[HTML]{FFD1D1}62.83 & \cellcolor[HTML]{FFD1D1}62.44 & \cellcolor[HTML]{E9FFC0}68.27 \\
MCC         & \cellcolor[HTML]{FFD1D1}62.36 & \cellcolor[HTML]{E9FFC0}69.65 & \cellcolor[HTML]{E9FFC0}70.06 & \cellcolor[HTML]{E9FFC0}69.66 & \cellcolor[HTML]{E9FFC0}69.68 & \cellcolor[HTML]{FFD1D1}63.21 & \cellcolor[HTML]{FFD1D1}40.48 & \cellcolor[HTML]{FFD1D1}24.72 & \cellcolor[HTML]{E9FFC0}66.84 & \cellcolor[HTML]{FFD1D1}55.12 & \cellcolor[HTML]{FFD1D1}54.66 & \cellcolor[HTML]{FFD1D1}35.13 & \cellcolor[HTML]{E9FFC0}66.79 & \cellcolor[HTML]{E9FFC0}65.28 & \cellcolor[HTML]{E9FFC0}69.11 & \cellcolor[HTML]{E9FFC0}72.41 \\
MCD         & \cellcolor[HTML]{FFD1D1}60.80 & \cellcolor[HTML]{FFD1D1}60.31 & \cellcolor[HTML]{FFD1D1}46.45 & \cellcolor[HTML]{FFD1D1}60.26 & \cellcolor[HTML]{FFD1D1}31.23 & \cellcolor[HTML]{FFD1D1}16.54 & \cellcolor[HTML]{FFD1D1}28.58 & \cellcolor[HTML]{FFD1D1}8.99  & \cellcolor[HTML]{E9FFC0}63.83 & \cellcolor[HTML]{FFD1D1}51.06 & \cellcolor[HTML]{FFD1D1}47.44 & \cellcolor[HTML]{FFD1D1}13.66 & \cellcolor[HTML]{E9FFC0}64.43 & \cellcolor[HTML]{FFD1D1}56.44 & \cellcolor[HTML]{E9FFC0}63.83 & \cellcolor[HTML]{E9FFC0}67.75 \\
MMD         & \cellcolor[HTML]{FFD1D1}60.37 & \cellcolor[HTML]{E9FFC0}65.98 & \cellcolor[HTML]{E9FFC0}63.56 & \cellcolor[HTML]{E9FFC0}66.00 & \cellcolor[HTML]{E9FFC0}63.56 & \cellcolor[HTML]{FFD1D1}54.83 & \cellcolor[HTML]{FFD1D1}51.93 & \cellcolor[HTML]{FFD1D1}35.22 & \cellcolor[HTML]{FFD1D1}61.37 & \cellcolor[HTML]{FFD1D1}46.66 & \cellcolor[HTML]{FFD1D1}58.41 & \cellcolor[HTML]{FFD1D1}40.08 & \cellcolor[HTML]{FFD1D1}61.57 & \cellcolor[HTML]{FFD1D1}61.06 & \cellcolor[HTML]{E9FFC0}63.83 & \cellcolor[HTML]{E9FFC0}67.44 \\ \hline
Avg.        & \cellcolor[HTML]{FFD1D1}60.86 & \cellcolor[HTML]{E9FFC0}66.29 & \cellcolor[HTML]{E9FFC0}63.60 & \cellcolor[HTML]{E9FFC0}66.26 & \cellcolor[HTML]{FFD1D1}60.81 & \cellcolor[HTML]{FFD1D1}49.96 & \cellcolor[HTML]{FFD1D1}46.00 & \cellcolor[HTML]{FFD1D1}25.91 & \cellcolor[HTML]{E9FFC0}64.32 & \cellcolor[HTML]{FFD1D1}51.59 & \cellcolor[HTML]{FFD1D1}56.33 & \cellcolor[HTML]{FFD1D1}34.54 & \cellcolor[HTML]{E9FFC0}64.53 & \cellcolor[HTML]{FFD1D1}62.14 & \cellcolor[HTML]{E9FFC0}65.55 & \cellcolor[HTML]{E9FFC0}69.87 \\
Avg. Rank   & 8.50                          & 3.33                          & 3.92                          & 3.00                          & 4.42                          & 11.00                         & 12.33                         & 15.00                         & 6.00                          & 11.33                         & 10.33                         & 14.00                         & 5.17                          & 7.33                          & 4.33                          & -                             \\
Correlation & 0.29                          & 0.62                          & 0.62                          & 0.65                          & 0.58                          & 0.01                          & -0.40                         & -0.60                         & 0.35                          & 0.30                          & -0.47                         & -0.15                         & 0.36                          & 0.30                          & 0.50                          & -                             \\ \midrule
Source-only & \cellcolor[HTML]{E9FFC0}63.58 & \cellcolor[HTML]{FFD1D1}49.81 & \cellcolor[HTML]{FFD1D1}48.04 & \cellcolor[HTML]{FFD1D1}48.02 & \cellcolor[HTML]{FFD1D1}48.04 & \cellcolor[HTML]{FFD1D1}47.88 & \cellcolor[HTML]{FFD1D1}32.62 & \cellcolor[HTML]{FFD1D1}46.35 & \cellcolor[HTML]{FFD1D1}54.90 & \cellcolor[HTML]{E9FFC0}63.69 & \cellcolor[HTML]{FFD1D1}49.31 & \cellcolor[HTML]{FFD1D1}32.49 & \cellcolor[HTML]{FFD1D1}49.30 & \cellcolor[HTML]{FFD1D1}49.22 & \cellcolor[HTML]{C0C0C0}63.48 & \cellcolor[HTML]{E9FFC0}65.60 \\
\bottomrule
\end{tabular}
}
\end{table*}

\keypoint{Adaptation Algorithms \& Validators} We consider six representative domain adaptation algorithms, spanning both recent and classic methods and a variety of underlying principles. These include the pseudo-label based \textbf{ATDOC} \cite{Liang2021DomainClassifier}; domain-adversarial learning with the seminal \textbf{DANN} \cite{ganin2016domainAdversarialNN}; domain-alignment with \textbf{MMD} \cite{Long2015LearningNetworks}; \textbf{BNM} and \textbf{MCC} which optimise the target label distribution under nuclear norm prior and minimum class confusion priors respectively, and the classifier-discrepancy-based \textbf{MCD} \cite{Saito2018MaximumAdaptation}. We explore tuning these models with a large number of potential validation criteria as listed in  \Cref{tab:validators}.

We start by finetuning a network on the source task. We take ResNet50 weights pretrained on ImageNet \cite{He2016DeepRecognition} for all datasets apart from MNIST-M where a smaller CNN is used. The final classification layer is replaced by an MLP head consisting of two blocks of \{Linear, ReLU, Dropout\} followed by a final linear layer. We finetune only this head on the source task using a standard categorical cross-entropy loss. 10 models are trained with learning rates sampled uniformly at random from a logarithmic scale between $10^{-5} - 10^{-1}$. These runs form the set of checkpoints for the \textit{source-only} model. \doublecheck{They are not used for evaluating the validation criteria, but we report performances at times for comparison.} 
When training each adaptation algorithm, we use the source-only model weights as initialisation for both the backbone and MLP head. The specific source-only checkpoint used as initialisation is the one with the highest source validation accuracy and in case of ties we select the checkpoint trained for the fewest amount of epochs. 

For each of our 6 adaptation algorithms, we sample 10 sets of hyperparameters and train one model per set. The training uses both source and target data. The number of epochs depends on the dataset and specific target domain, but in all cases, we save 20 checkpoints during the course of training. The optimizer is Adam with parameters \texttt{\{betas=(0.9, 0.999)\}} and the weight decay is always 0.0001. The learning rate is always part of the sampled hyperparameters, and it is updated during training via cosine annealing with a warmup phase during the first 5\% of training. The full details of our training procedure and hyperparameter search spaces can be found in \cref{sec:training_details}.

\cut{
\begin{table*}
\caption{Comparison of validation criteria for model selection in UDA. Averages over all 21 domain transfers evaluated. We report (i) the target test performance for the top models selected by each validator, and (ii) the correlation coefficient between the validator scores and the test performance over all hyperparameters and checkpoints.}
  \label{tab:uda}
\resizebox{1.0\linewidth}{!}{
  \begin{tabular}{l|ccccccccccccccc|c}
\toprule
{} &  RankMe  &  AMI  &  ARI  &  V-Measure  &  FMI  &  Silhouette  &  DBI  &  CHI  &  BNM &  MMD &  CORAL &  SND &  IM &  Entropy &  Accuracy  & Oracle \\
\midrule
ATDOC&58.24 &67.70 &67.71 &67.79 &67.71 &46.73 &49.55 &16.55 &64.29 &52.46 &55.96 & 24.13 & 64.61 &61.23 &  68.06 &   72.24 \\
BNM &61.36 &69.32 &69.48 &69.29 &69.48 &62.42 &51.58 &33.10 &66.88 &52.64 &60.92 & 47.70 & 67.01 &65.98 &  66.02 &   71.09 \\
DANN&62.00 &64.76 &64.35 &64.55 &63.23 &56.06 &53.86 &36.89 &62.72 &51.62 &60.61 & 46.51 & 62.79 &62.83 &  62.44 &   68.27 \\
MCC &62.36 &69.65 &70.06 &69.66 &69.68 &63.21 &40.48 &24.72 &66.84 &55.12 &54.66 & 35.13 & 66.79 &65.28 &  69.11 &   72.41 \\
MCD &60.80 &60.31 &46.45 &60.26 &31.23 &16.54 &28.58 & 8.99 &63.83 &51.06 &47.44 & 13.66 & 64.43 &56.44 &  63.83 &   67.75 \\
MMD &60.37 &65.98 &63.56 &66.00 &63.56 &54.83 &51.93 &35.22 &61.37 &46.66 &58.41 & 40.08 & 61.57 &61.06 &  63.83 &   67.44 \\
\midrule
Avg.&60.86 &66.29 &63.60 &66.26 &60.81 &49.96 &46.00 &25.91 &64.32 &51.59 &56.33 & 34.54 & 64.53 &62.14 &  65.55 &   69.87 \\
Avg. Rank   & 8.50 & 3.33 & 3.92 & 3.00 & 4.42 &11.00 &12.33 &15.00 & 6.00 &11.33 &10.33 & 14.00 &5.17 &7.33 &   4.33 &- \\
Correlation & 0.29 & 0.62 & 0.62 & 0.65 & 0.58 & 0.01 & -0.40 & -0.60 & 0.35 & 0.30 & -0.47 & -0.15 & 0.36 & 0.30 & 0.50 & - \\
\midrule
Source-only &  63.58 &49.81 &48.04 &  48.02 &48.04 & 47.88 &32.62 &46.35 &  54.90 & 63.69 &  49.31 & 32.49 & 49.30 &  49.22 &   63.48 &   65.60 \\
\bottomrule
  \end{tabular}
}
\end{table*}
}

\subsubsection{Results} The results in \Cref{tab:uda} report the performance of each adaptation algorithm and validation criterion combination, averaged over all 21 domain transfer tasks -- in terms of both test accuracy after HPO and the weighted Spearman correlation coefficient between validation scores and testing accuracy. More detailed correlation plots are given in \Cref{sec:correlations}. \Cref{tab:percentage_over_baseline} shows how easily tunable the algorithms are, via the percentage of all algorithm checkpoints outperforming the baseline.

\vspace{0.1cm}\noindent\textbf{How well does unsupervised validation work?}\quad
From  \Cref{tab:uda} we can draw a rich set of observations: (1) The validation criteria have varying ability to predict the test accuracy and thus drive HPO in domain adaptation. This is visible in the correlation scores, ranging from -.60 to .65 correlation coefficient at best; and the significant variability of testing performance when using the various criteria to drive HPO. Importantly, it is also visible in the gap between the performance of the best criteria and the best-case oracle criterion. (2) The best validation criterion is the previously un-studied V-measure score, which has the best average rank of 3.0 across all the validators, and closes 70\% of the gap between the baseline of 63.5\% and oracle upper bound of 72.4\% when paired with the MCC adapter. 
%(69.7-63.5)/(72.4-63.5)
(3) The DA algorithms themselves vary substantially in how easy they are to tune, with MCD and ATDOC for example being highly dependent on choice of validator, versus BNM which is comparatively insensitive to the choice of validator. Practitioners may prefer to opt for comparatively easy to tune adaptation algorithms, given the challenge of validation for DA.
\cut{(4) A final novel observation is the importance of source model selection (bottom row), which is not usually considered in DA. A strong source checkpoint, for example, as chosen by the RankMe validator, performs surprisingly close to many of the actual adaptation algorithms. This suggests that simply carefully controlling how much opportunity the model has to specialise to the source data can be a very strong baseline, in line with \cite{li2022finding}.}

%\subsubsection{How well do validation criteria correlate with testing performance?} Using the checkpoints of all our trained models, we compute the true target domain testing performance for each; as well as each checkpoint's score under the various validation criteria. The results, aggregated across all pairs of source and target domains in the Office-31 dataset, are shown in Figure~\ref{fig:uda}.
%From the results, we can see that our G-score and H-score have the highest correlation with target test accuracy.

\begin{table}[t]
    \centering
    \caption{\rebuttal{Percentage of all algorithm checkpoints which outperform the baseline source-only model, in the UDA setting. The ATDOC, BNM and MCC algorithms are the most easily tunable, with over 40\% of hyperparameter choices leading to better-performing models. We also see that when selecting checkpoints with the oracle validator, 18.3\% of the source-only checkpoints outperform the one selected as the baseline using the source validation accuracy validator.}}
    \label{tab:percentage_over_baseline}
    \rebuttal{
    \resizebox{0.6\linewidth}{!}{
    \begin{tabular}{ccccccc}
        \toprule
        Source-only & ATDOC & BNM & DANN & MCC & MCD & MMD \\
        \midrule
        18.3 & 42.8 & 48.4 & 24.9 & 43.4 & 14.6 & 29.5 \\
        \bottomrule
    \end{tabular}
    }
    }
\end{table}

\begin{figure}[b]
\centering
\includegraphics[width=1.0\linewidth]{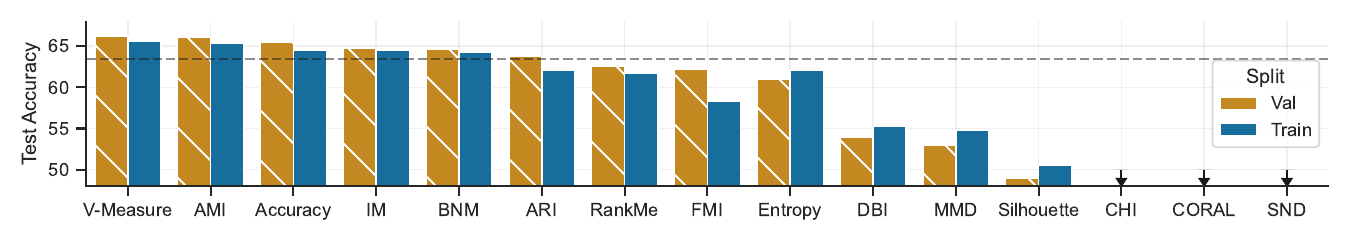}
\caption{Comparison of split for evaluation of validation criteria. We report the average target test accuracy of selected models for each validator when applied on (blue) target train data and (orange) target validation data. The dashed line is the source-only model performance.}
\label{fig:train_vs_val}
\end{figure}

%\subsubsection{Which validation criterion leads to the best generalisation performance when used for model selection?} We next use the various scores to perform model selection and hyperparameter optimisation for each of the base DA algorithms.
%From the results in Table~\ref{tab:uda}, we can see that our proposed G-score and H-score lead to the highest average accuracy and lowest average rank across all the DA algorithms. This confirms that our scores are not only highly correlated with testing accuracy, but effective for use in model selection and HPO.

\cut{
\begin{table*}[t]
  \centering
  \caption{Comparison of split for evaluation of validation criteria. We report the average target test accuracy of selected models for each validator when applied on (1) target train data and (2) target validation data.}
  \label{tab:trainOrValSplit}
\resizebox{1.0\linewidth}{!}{
\begin{tabular}{l|ccccccccccccccc}
\toprule
{} &  RankMe &  AMI &  ARI &  V-Measure &  FMI &  Silhouette &  DBI &  CHI &  BNM &  MMD &  CORAL &  SND &  IM &  Entropy &  Accuracy \\
\midrule
Train & 61.77 & 65.43 & 62.15 & 65.59 & 58.27 & 50.56 & 55.23 & 29.19 & 64.32 & 54.76 & 28.97 & 24.12 & 64.53 &  62.14 &  64.50 \\
Val& 62.54 & 66.13 & 63.78 & 66.22 & 62.27 & 48.90 & 53.97 & 26.31 & 64.66 & 53.01 & 28.88 & 23.07 & 64.82 &  60.98 &  65.55 \\
\bottomrule
\end{tabular}
}
\end{table*}
}

\noindent\textbf{What is the impact of validating on training vs validation splits?}\quad An important design choice in validation is which data split the validator is evaluated on. 
\label{sec:train_vs_val} As discussed in \cite{Musgrave2020UnsupervisedCheck,musgrave2022benchmarkingUDA}, while prior work that validates on the source domain has fairly consistently used the source validation set; prior work that validates on the unlabelled target domain has been inconsistent in the choice between validating using the train or an independent val split. Since learning is driven by applying an adaptation loss on the target train set, there is the possibility of overfitting during unsupervised adaptation. Thus we conjecture that one should validate on a disjoint split of the target domain. \doublecheck{In \cref{tab:uda}, we avoided this issue by taking the best split for each validator.}
We now analyse this issue by comparing using the train vs validation split for evaluating criteria. From the results in \cref{fig:train_vs_val} we see that for all the top-performing criteria the val split is preferred. While this result might seem unsurprising in retrospect, we emphasise that the use of a val split is NOT standard practice in the literature, even in thorough recent evaluations \cite{musgrave2022benchmarkingUDA}. We show that there is thus a trade-off between using held-out validation data to improve the validators, and using as much training data as possible to improve adaptation.

\begin{table*}[t]
  \centering
  \caption{Comparison of validation criteria for model selection in SFDA on Office-Home. We report (i) the target test performance for the top models selected by each validator, and (ii) the correlation coefficient between the validator scores and the test performance over all hyperparameters and checkpoints. For each algorithm, we include the source-only checkpoint in the pool available to validators (indicated by the ``+SO'' suffix). The colour of a cell indicates whether that model/validator combination beats the source-only model (green) or not (red), with a darker red colour meaning it fails to achieve half of the source-only model performance.}
  \label{tab:sfda}

  \resizebox{1.0\linewidth}{!}{
  \begin{tabular}{l|ccccccccccccc|c}
  \toprule
  {}        & RankMe                        & AMI                           & ARI                           & V-Measure                     & FMI                           & Silhouette                    & DBI                           & CHI                           & BNM                           & SND                           & IM                            & Entropy                       & Accuracy                      & Oracle                        \\
\midrule
AAD+SO      & \cellcolor[HTML]{E9FFC0}61.63 & \cellcolor[HTML]{E9FFC0}57.41 & \cellcolor[HTML]{FFA6A6}1.60  & \cellcolor[HTML]{E9FFC0}60.19 & \cellcolor[HTML]{FFA6A6}1.59  & \cellcolor[HTML]{FFA6A6}1.74  & \cellcolor[HTML]{FFD1D1}53.62 & \cellcolor[HTML]{FFA6A6}1.90  & \cellcolor[HTML]{E9FFC0}62.51 & \cellcolor[HTML]{FFD1D1}56.29 & \cellcolor[HTML]{E9FFC0}59.40 & \cellcolor[HTML]{FFA6A6}4.78  & -                             & \cellcolor[HTML]{E9FFC0}65.71 \\
NRC+SO      & \cellcolor[HTML]{E9FFC0}57.30 & \cellcolor[HTML]{E9FFC0}58.44 & \cellcolor[HTML]{FFA6A6}6.74  & \cellcolor[HTML]{E9FFC0}62.73 & \cellcolor[HTML]{FFA6A6}6.70  & \cellcolor[HTML]{FFA6A6}1.62  & \cellcolor[HTML]{FFD1D1}45.12 & \cellcolor[HTML]{FFA6A6}1.70  & \cellcolor[HTML]{E9FFC0}58.18 & \cellcolor[HTML]{E9FFC0}56.84 & \cellcolor[HTML]{E9FFC0}57.71 & \cellcolor[HTML]{FFD1D1}40.70 & -                             & \cellcolor[HTML]{E9FFC0}64.93 \\
SHOT+SO     & \cellcolor[HTML]{E9FFC0}59.20 & \cellcolor[HTML]{E9FFC0}59.73 & \cellcolor[HTML]{E9FFC0}60.88 & \cellcolor[HTML]{E9FFC0}60.99 & \cellcolor[HTML]{E9FFC0}59.38 & \cellcolor[HTML]{E9FFC0}57.54 & \cellcolor[HTML]{FFD1D1}55.41 & \cellcolor[HTML]{FFD1D1}46.72 & \cellcolor[HTML]{FFD1D1}54.14 & \cellcolor[HTML]{E9FFC0}57.13 & \cellcolor[HTML]{FFD1D1}54.52 & \cellcolor[HTML]{E9FFC0}59.51 & -                             & \cellcolor[HTML]{E9FFC0}64.04 \\
\midrule
Avg.        & \cellcolor[HTML]{E9FFC0}59.38 & \cellcolor[HTML]{E9FFC0}58.52 & \cellcolor[HTML]{FFA6A6}23.07 & \cellcolor[HTML]{E9FFC0}61.30 & \cellcolor[HTML]{FFA6A6}22.56 & \cellcolor[HTML]{FFA6A6}20.30 & \cellcolor[HTML]{FFD1D1}51.38 & \cellcolor[HTML]{FFA6A6}16.77 & \cellcolor[HTML]{E9FFC0}58.28 & \cellcolor[HTML]{E9FFC0}56.75 & \cellcolor[HTML]{E9FFC0}57.21 & \cellcolor[HTML]{FFD1D1}35.00 & -                             & \cellcolor[HTML]{E9FFC0}64.89 \\
Avg. Rank   & 4.33                          & 3.33                          & 7.33                          & 1.67                          & 9.00                          & 9.67                          & 7.67                          & 10.67                         & 5.00                          & 6.67                          & 6.00                          & 6.67                          & -                             & -                             \\
Correlation & -0.02                         & 0.11                          & -0.32                         & 0.09                          & -0.08                         & -0.54                         & 0.01                          & -0.77                         & -0.01                         & 0.06                          & 0.02                          & -0.11                         & -                             & -                             \\
\midrule
Source-only & -                             & -                             & -                             & -                             & -                             & -                             & -                             & -                             & -                             & -                             & -                             & -                             & \cellcolor[HTML]{C0C0C0}56.49 & -                             \\
  \bottomrule
  \end{tabular}
  }
\end{table*}

% target_train_preds_rank_me_score - RankMe (Target Train Preds): 
% target_train_class_ami_score - AMI (Target Train Features): 
% target_val_logits_class_ari_score - ARI (Target Val Logits): 
% target_train_class_v_measure_score - V-Measure (Target Train Features): 
% target_train_class_fmi_score - FMI (Target Train Features): 
% target_train_logits_class_silhouette_score - Silhouette (Target Train Logits): 
% target_train_logits_class_dbi_score - BDI (Target Train Logits): 
% target_train_class_chi_score - CHI (Target Train Features): 
% target_train_bnm_score - BNM (Target Train): 
% target_val_neg_snd_score - -SND (Target Val): 
% target_train_im_score - IM (Target Train): 
% target_train_entropy_score - Entropy (Target Train): 

\subsection{Source-free Domain Adaptation}
\subsubsection{Setup}
For SFDA we use the Office-Home dataset as a benchmark, covering all 12 domain shifts. The same source-only models that we produced for UDA are also used here for initialisation of the same architecture. Three recent SFDA algorithms adapt the model on target domain data, AAD \cite{yang2022attracting}, NRC \cite{yang2021exploiting} and SHOT \cite{liang2020sourceFree}. For each algorithm, we sample 10 sets of hyperparameters and train for 200 epochs. The setup follows the UDA setting described above, with the main difference being the adaptation algorithms and validators only have access to target data (\cref{fig:domain_split_and_use_diagram}). 
%
%\keypoint{Validators}
As the source domain is not available in this setting, we can only apply our validators to the target domain splits. This means CORAL and MMD are not applicable, since they need both domains to compute their scores. Following our results in Section \ref{sec:train_vs_val} we use the target validation split for all validators as the source data is absent in this case.

\subsubsection{Results} Analogous to UDA, the results in \Cref{tab:sfda} report the performance of each adaptation algorithm and validation criterion combination, averaged over all 12 domain transfer SFDA tasks – in terms of both test accuracy after HPO and the weighted Spearman correlation coefficient between validation scores and testing accuracy. More detailed correlation plots are given in Appendix E.

\ques{How does unsupervised validation work in the absence of source data?}
From the results in \Cref{tab:sfda} we can draw a set of conclusions analogously to UDA. Specifically, (i) Here, the best validators are RankMe and V-measure, with V-measure closing up to 75\% of the gap between the baseline and oracle when combined with NRC.
%(64.9-62.7)/(64.9-56.5)
%However, the validator correlations are now \emph{much} weaker than in \Cref{tab:uda}, reaching 0.11 at best for AMI. There is a substantial gap of 5\% on average between the best validator and the oracle, indicating that more work is necessary on validation criteria. 
(ii) However the AAD and NRC algorithms are highly sensitive to validator choice, with the weaker validators such as FMI and CHI producing catastrophically poor performance, suggesting that SHOT might be preferred in practice even though NRC has the best accuracy when paired with its preferred validator, and AAD when validated with the oracle. 
(iii) Many algorithm-validator combinations lead to \emph{worse} performance than the baseline source-only model. This highlights an important point that in the absence of highly reliable validation criteria, DA algorithms pose a risk of making the performance even worse. This issue is one which is not widely analysed in academic DA but is obviously crucial. Please note that we also included the model initialization (i.e. the source-only model) as one of the checkpoints available for selection by the criteria. However, many validators 
%such as ARI, FMI, Silhuette, DBI, CHI, BNM and Entropy 
fail to detect adaptation failure and select a safe source-only model.

% (iii) \doublecheck{A potential remedy to this issue is to include the source-only initialisation checkpoint in the pool available to validators and hope that when the adaptor fails, the validator is able to resort back to the source-only weights as a fail-safe option. We can see in the lower part of \cref{tab:sfda} that this strategy offers some improvements but often still results in worse performance. This highlights that sometimes adaptation may be riskier than simply applying a baseline source model.}

%\keypoint{Questions}

\begin{table*}[t]
\centering
\caption{Test-Time Adaptation on CIFAR10-C, at corruption level 5. We use the episodic setup where the model is reset after each batch. For each algorithm, we include the source-only checkpoint in the pool available to validators (indicated by the ``+SO'' suffix). The colour of a cell indicates whether that model/validator combination beats the source-only model (green) or not (red).}% Only the target domain is available and it is used both for adaptation, validation and evaluating performance.}
\label{tab:tta}

  \resizebox{1.0\linewidth}{!}{
\begin{tabular}{l|ccccccccccccc|c}
\toprule
{} &  RankMe &  AMI &  ARI &  V-Measure &  FMI &  Silhouette &  DBI &  CHI &  BNM &  SND &  IM &  Entropy & Accuracy & Oracle \\
\midrule
SHOT+SO     & \cellcolor[HTML]{E9FFC0}78.38 & \cellcolor[HTML]{FFD1D1}37.37 & \cellcolor[HTML]{FFD1D1}36.85 & \cellcolor[HTML]{FFD1D1}37.69 & \cellcolor[HTML]{FFD1D1}36.86 & \cellcolor[HTML]{FFD1D1}36.50 & \cellcolor[HTML]{FFD1D1}38.60 & \cellcolor[HTML]{FFD1D1}43.87 & \cellcolor[HTML]{FFD1D1}45.22 & \cellcolor[HTML]{FFD1D1}53.48 & \cellcolor[HTML]{FFD1D1}46.80 & \cellcolor[HTML]{FFD1D1}39.14 & -                             & \cellcolor[HTML]{E9FFC0}86.24 \\
TENT+SO     & \cellcolor[HTML]{E9FFC0}84.06 & \cellcolor[HTML]{E9FFC0}84.14 & \cellcolor[HTML]{E9FFC0}84.17 & \cellcolor[HTML]{E9FFC0}84.13 & \cellcolor[HTML]{E9FFC0}84.17 & \cellcolor[HTML]{E9FFC0}84.23 & \cellcolor[HTML]{E9FFC0}81.84 & \cellcolor[HTML]{E9FFC0}75.98 & \cellcolor[HTML]{E9FFC0}84.22 & \cellcolor[HTML]{E9FFC0}79.89 & \cellcolor[HTML]{E9FFC0}84.21 & \cellcolor[HTML]{E9FFC0}83.20 & -                             & \cellcolor[HTML]{E9FFC0}84.76 \\
\midrule
Avg.        & \cellcolor[HTML]{E9FFC0}81.22 & \cellcolor[HTML]{FFD1D1}60.76 & \cellcolor[HTML]{FFD1D1}60.51 & \cellcolor[HTML]{FFD1D1}60.91 & \cellcolor[HTML]{FFD1D1}60.51 & \cellcolor[HTML]{FFD1D1}60.36 & \cellcolor[HTML]{FFD1D1}60.22 & \cellcolor[HTML]{FFD1D1}59.93 & \cellcolor[HTML]{FFD1D1}64.72 & \cellcolor[HTML]{FFD1D1}66.68 & \cellcolor[HTML]{FFD1D1}65.51 & \cellcolor[HTML]{FFD1D1}61.17 & -                             & \cellcolor[HTML]{E9FFC0}85.50 \\
Avg. Rank   & 4.50                          & 7.50                          & 7.75                          & 7.50                          & 7.25                          & 6.50                          & 8.50                          & 8.50                          & 3.00                          & 6.50                          & 3.00                          & 7.50                          & -                             & -                             \\
Correlation & 0.10                          & 0.13                          & 0.14                          & 0.13                          & 0.14                          & 0.13                          & -0.56                         & 0.09                          & 0.09                          & -0.55                         & 0.08                          & -0.02                         & -                             & -                             \\ \hline
Source-only & -                             & -                             & -                             & -                             & -                             & -                             & -                             & -                             & -                             & -                             & -                             & -                             & \cellcolor[HTML]{C0C0C0}70.64 & -                             \\
\bottomrule
\end{tabular}
  }
\end{table*}

\subsection{Test-Time Adaptation}
\subsubsection{Setup}
We next adopt the TTA setting, where a pre-trained model adapts to the test data as it comes, one batch at a time. For simplicity, we use the episodic setting \cite{wang2021tent} where the model is reset after each batch.
% \keypoint{Datasets}
We use the most common TTA benchmark of CIFAR10-C, consisting of 15 versions of the CIFAR10 test set with various corruptions applied, including Gaussian noise, pixelation and fog. Additionally, we investigate whether existing TTA algorithms are able to deal with the more complex distribution shifts from Office-Home, using all 12 domain shift setups.
% \keypoint{Algorithms}
For CIFAR10-C, we use the pre-trained CIFAR10 checkpoint of \cite{ttt_plus_plus} as our source-only model and initialisation for the TTA algorithms. For Office-Home, we use the same source-only checkpoints as in the UDA and SFDA sections above. Two algorithms are trained: SHOT \cite{liang2020sourceFree} which uses information maximisation and pseudo-labelling to align target representations and TENT \cite{wang2021tent} which adapts by minimising the entropy of its predictions on the test batch.
% \keypoint{Validators}
As this setting only exposes a single batch to the model at a time, both training and validation use the same data. As in the SFDA setting, CORAL and MMD are not applicable, since they need both domains to compute their scores. This also means that there is no validator based on accuracy, as it requires source data to be computed.

% \keypoint{Questions}

\subsubsection{Results} Tables~\ref{tab:tta} and \ref{tab:tta_officehome} report the performance of each adaptation algorithm and validation criterion combination averaged over all CIFAR-C and Office-Home TTA tasks. 

\ques{Is Test-Time Adaptation effective when performing proper model selection?}
The results for CIFAR in Table~\ref{tab:tta} lead to a different conclusion from that of UDA and SFDA. (i) RankMe is again the best validation criterion, and interestingly, we see that the top validators now manage to almost match the oracle performance. (ii) TENT is robust to the choice of validator, with consistently good performance close to oracle. SHOT obtains reasonable performance only when RankMe validator is used. 
%This highlights that a good TTA performance highly relies on selecting a robust TTA method. 
%(iii) However, the correlation scores are very low for all methods, with a max correlation of 0.14. \textcolor{red}{Thus the success of TENT is attributed to "all checkpoints being quite effective" more than validator efficacy.}

While TENT-based TTA plus various validators above show a success case for good practice adaptation on CIFAR10-C, we next ask whether these good results persist to a real rather than synthetic adaptation task. Table~\ref{tab:tta_officehome}, shows the results of the  Office-Home benchmark. From the results, we can see that: (1) There is only a 2-3\% gap between the oracle best case and the baseline, suggesting that all algorithms struggle on this benchmark, even for best-case HPO. (2) Almost all algorithm-validator combinations are worse than the 57\% source-only accuracy, similar to the SFDA case discussed earlier. When we compare the adaptation performance with- and without- access to the source-only model in the pool of checkpoints for HPO, SHOT has little improvement. The validators are not able to respond to the destructive adaptation and fail to pick a safe pre-adaptation model. Thus, we suggest that the strong success of TTA methods on synthetic benchmarks may not be representative of real-world adaptation problems, especially when required to use fair validation.

%first see that TENT does not work in this case. And we surprisingly find that all validators fail to search for a good TTA model in this setting -- namely, results are all worse than oracle and even that of the source-only baseline, consistently. This can also be found in the demonstrated weaker correlations compared with above.
%Conclusively, we can see that under a strong domain shift, TTA methods may not work as is or due to a lack of good validators. 

% In this setting, the top validators manage to almost match the oracle performance, indicating that ... However, it is worth noting that this is an artificially constructed benchmark. We next investigate how TTA performs on a more realistic set of domain shifts from Office-Home.

\cut{
\begin{table}[t]
    \centering
    \caption{\rebuttal{Percentage of top-1 checkpoints, as selected by each validator, which outperform the baseline source-only model, in the UDA setting. The AMI, ARI, V-Measure and source validation accuracy are the most reliable, with 5 out of 6 algorithm choices leading to better-performing models.}}
    \label{tab:percentage_over_baseline}
    \resizebox{1.0\linewidth}{!}{
    \rebuttal{
    \begin{tabular}{ccccccccccccccc|c}
        \toprule
        RankMe  &  AMI  &  ARI  &  V-Measure  &  FMI  &  Silhouette  &  DBI  &  CHI  &  BNM &  MMD &  CORAL &  SND &  IM &  Entropy &  Accuracy \\
        \midrule
        0\%  &  83.3\%  &  83.3\%  &  83.3\%  &  66.7\%  &  0\%  &  0\%  &  0\%  &  66.7\% &  0\% &  0\% &  0\% &  66.7\% &  33.3\% &  83.3\% \\
        \bottomrule
    \end{tabular}
    }
    }
\end{table}
}

\begin{table*}[t]
  \centering
  \caption{Test-Time Adaptation on Office-Home. We use the episodic setup where the model is reset after each batch. Algorithms that include the source-only checkpoint in the pool available to validators are marked by the suffix ``+SO''. The colour of a cell indicates whether that model/validator combination beats the source-only model (green) or not (red), with a darker red colour meaning it fails to achieve half of the source-only model performance.}% Only the target domain is available and it is used both for adaptation, validation and evaluating performance.}
  \label{tab:tta_officehome}

  \resizebox{1.0\linewidth}{!}{
    \begin{tabular}{l|ccccccccccccc|c}
    \toprule
  {}        & RankMe                        & AMI                           & ARI                           & V-Measure                     & FMI                           & Silhouette                    & DBI                           & CHI                           & BNM                           & SND                           & IM                            & Entropy                      & Accuracy                      & Oracle                        \\
\midrule
SHOT        & \cellcolor[HTML]{FFA6A6}20.46 & \cellcolor[HTML]{FFA6A6}9.38  & \cellcolor[HTML]{FFA6A6}10.09 & \cellcolor[HTML]{FFA6A6}9.43  & \cellcolor[HTML]{FFA6A6}10.09 & \cellcolor[HTML]{FFA6A6}14.07 & \cellcolor[HTML]{FFD1D1}34.67 & \cellcolor[HTML]{FFA6A6}12.06 & \cellcolor[HTML]{FFA6A6}9.10  & \cellcolor[HTML]{FFD1D1}35.58 & \cellcolor[HTML]{FFA6A6}10.13 & \cellcolor[HTML]{FFA6A6}7.52 & -                             & \cellcolor[HTML]{E9FFC0}59.05 \\
TENT        & \cellcolor[HTML]{FFD1D1}42.40 & \cellcolor[HTML]{FFD1D1}37.57 & \cellcolor[HTML]{FFD1D1}42.79 & \cellcolor[HTML]{FFD1D1}43.49 & \cellcolor[HTML]{FFD1D1}40.77 & \cellcolor[HTML]{FFA6A6}2.25  & \cellcolor[HTML]{FFD1D1}39.90 & \cellcolor[HTML]{FFA6A6}3.61  & \cellcolor[HTML]{FFD1D1}44.93 & \cellcolor[HTML]{FFD1D1}39.20 & \cellcolor[HTML]{FFD1D1}44.24 & \cellcolor[HTML]{FFA6A6}2.37 & -                             & \cellcolor[HTML]{FFD1D1}49.28                         \\
\midrule
SHOT+SO     & \cellcolor[HTML]{FFA6A6}20.46 & \cellcolor[HTML]{FFA6A6}9.32  & \cellcolor[HTML]{FFA6A6}12.40 & \cellcolor[HTML]{FFA6A6}9.12  & \cellcolor[HTML]{FFA6A6}12.05 & \cellcolor[HTML]{FFA6A6}14.84 & \cellcolor[HTML]{FFD1D1}34.67 & \cellcolor[HTML]{FFA6A6}11.63 & \cellcolor[HTML]{FFA6A6}9.10  & \cellcolor[HTML]{FFD1D1}35.58 & \cellcolor[HTML]{FFA6A6}10.81 & \cellcolor[HTML]{FFA6A6}7.54 & -                             & \cellcolor[HTML]{E9FFC0}60.20 \\
TENT+SO     & \cellcolor[HTML]{FFD1D1}42.59 & \cellcolor[HTML]{FFD1D1}55.91 & \cellcolor[HTML]{FFD1D1}55.91 & \cellcolor[HTML]{FFD1D1}55.91 & \cellcolor[HTML]{FFD1D1}55.91 & \cellcolor[HTML]{FFA6A6}6.35  & \cellcolor[HTML]{FFD1D1}39.90 & \cellcolor[HTML]{FFD1D1}45.09 & \cellcolor[HTML]{FFD1D1}46.01 & \cellcolor[HTML]{FFD1D1}39.20 & \cellcolor[HTML]{FFD1D1}45.75 & \cellcolor[HTML]{FFA6A6}2.37 & -                             & \cellcolor[HTML]{E9FFC0}57.15 \\
\midrule
Avg.        & \cellcolor[HTML]{FFD1D1}31.52 & \cellcolor[HTML]{FFD1D1}32.62 & \cellcolor[HTML]{FFD1D1}34.16 & \cellcolor[HTML]{FFD1D1}32.52 & \cellcolor[HTML]{FFD1D1}33.98 & \cellcolor[HTML]{FFA6A6}10.60 & \cellcolor[HTML]{FFD1D1}37.29 & \cellcolor[HTML]{FFA6A6}28.36 & \cellcolor[HTML]{FFA6A6}27.56 & \cellcolor[HTML]{FFD1D1}37.39 & \cellcolor[HTML]{FFA6A6}28.28 & \cellcolor[HTML]{FFA6A6}4.95 & -                             & \cellcolor[HTML]{E9FFC0}58.68 \\
Avg. Rank   & 5.50                          & 5.75                          & 3.75                          & 6.25                          & 4.25                          & 7.50                          & 5.50                          & 7.00                          & 8.00                          & 5.50                          & 7.00                          & 12.00                        & -                             & -                             \\
Correlation & 0.28                          & 0.22                          & 0.25                          & 0.22                          & 0.29                          & -0.18                         & 0.11                          & -0.05                         & -0.02                         & -0.03                         & 0.02                          & -0.53                        & -                             & -                             \\ \hline
Source-only & -                             & -                             & -                             & -                             & -                             & -                             & -                             & -                             & -                             & -                             & -                             & -                            & \cellcolor[HTML]{C0C0C0}56.97 & -                             \\
    \bottomrule
    \end{tabular}
  }
\end{table*}
% validator versions for ++ adaptors:
%['target_train_logits_rank_me_score',
% 'target_train_class_ami_score',
% 'target_train_class_ari_score',
% 'target_train_class_v_measure_score',
% 'target_train_class_fmi_score',
% 'target_train_class_silhouette_score',
% 'target_train_class_dbi_score',
% 'target_train_class_chi_score',
% 'target_train_bnm_score',
% 'target_train_neg_snd_score',
% 'target_train_im_score',
% 'target_train_entropy_score',
% 'target_train_acc_score']

\section{Conclusion} \label{sec:conclusion} 
In this work, we performed a comprehensive study of HPO and model selection for domain adaptation, covering 10 algorithms and 15 validators across three settings (UDA, SFDA and TTA). We have found that previously unexplored validators like RankMe and V-Measure perform well across several settings, but the optimal validator is setting and algorithm dependent. Thus practitioners may wish to consider tuning sensitivity as a key factor for algorithm selection beyond reported performance on academic benchmarks. We have also highlighted some surprising results: (i) A strong source checkpoint can be competitive with UDA algorithms when using the RankMe or MMD validators. (ii) Even when validating among both source-only and algorithm checkpoints, performance may be worse than abandoning adaptation altogether and simply using a source-only model as selected by source accuracy -- especially in SFDA and TTA. This is a major risk and failure case that will preclude deployment of adaptation in real applications, and one which we encourage future academic work to study. (iii) While TTA algorithms have attracted attention for their strong performance on simple synthetic benchmarks, they fail on more complex distribution shifts such as Office-Home. Future work should carefully consider model selection pipelines, choice of validators, data splits and, at times, whether to perform adaptation at all.

\section{Limitations} \label{sec:limitations}
Since this paper covers multiple settings and many algorithms, we were only able to run each algorithm on each domain shift with 10 hyperparameter choices. Ideally, this number would be higher, but adapting many models is expensive, and in reality a practitioner would also face computational limitations. If we had been able to increase this number, the effects would likely be minor improvements in the highest scores achievable by the algorithms, but it would not guarantee that any given validator would choose a better checkpoint. 
A second limitation is our preliminary study of regression adaptation. Many validators are designed for classification and performance when applied to regression may suffer. However, we view this as a call for study of validation criteria across the full range of adaptation applications from regression to segmentation and detection. 
%of validators to the regression setting . More effort can be put into defining suitable validators for this task, but as this is such an unexplored area we present an initial investigation into a simple application of existing validators.

\section{Broader Impact Statement} \label{sec:impact}
As distribution shifts often occur in real-world problems, the use of domain adaptation techniques to tackle them has become commonplace. We have highlighted weaknesses in the standard practice found in DA literature which, if applied in safety-critical scenarios, could lead to catastrophic outcomes. We have therefore outlined a set of better practices for practitioners looking to apply DA to their problems and presented a more realistic view of the field's current potential. Our hope is that this will decrease the likelihood of bad outcomes in its real-world application. The environmental impact of running a large-scale benchmarking study is significant. However, we aim to provide a clear pipeline for future work to use, which can ultimately reduce unnecessary computation due to bad practice.

\newpage

{\small
\bibliography{refs}
}

% The 9 pages allocated for the main paper must include a broader impact
% statement regarding the approach, datasets and applications proposed/used in
% your paper. It should reflect on the environmental, ethical and societal
% implications of your work. The statement should require at most one page and
% must be included both at submission and camera-ready time.
%
% If authors have reflected on their work and determined that there are no
% likely negative broader impacts, they may use the following statement:
%
% After careful reflection, the authors have determined that this work presents
% no notable negative impacts to society or the environment.
%
% This section is included in the template as a default, but you can also place these
% discussions anywhere else in the main paper, e.g., in the introduction/future work.
%
% The Centre for the Governance of AI has written an excellent guide for writing
% good broader impact statements (for the NeurIPS conference) that may be a
% useful resource for AutoML-Conf authors:
%
% https://medium.com/@GovAI/a-guide-to-writing-the-neurips-impact-statement-4293b723f832
\cut{
\newpage
\section{Submission Checklist}

% The submission checklist is a combination of the NeurIPS '21 checklist:
%
%https://neurips.cc/Conferences/2021/PaperInformation/PaperChecklist
%
% and the NAS checklist:
%
%https://www.automl.org/wp-content/uploads/NAS/NAS_checklist.pdf
%
% For each question, change the default \answerTODO{} to either:
%
% \answerYes{[justification]},
% \answerNo{[justification]}, or
% \answerNA{[justification]}.
%
% *You must include a brief justification to your answer,* either by
% referencing the appropriate section of your paper or providing a brief inline
% description.  For example:
%
% - Did you include the license of the code and datasets?
%\answerYes{See Section~\ref{sec:code}.}
%
% - Did you include all the code for running experiments?
%\answerNo{We include the code we wrote, but it depends on proprietary
%libraries for executing on a compute cluster and as such will not be
%runnable without modifications. We also include a runnable sequential
%version of the code that we also report experiments in the paper with.}
%
% - Did you include the license of the datasets?
%\answerNA{Our experiments were conducted on publicly available datasets and
%we did not introduce new datasets.}
%
% Please note that if you answer a question with \answerNo{}, we expect that you
% compensate for it (e.g., if you cannot provide the full evaluation code, you
% should at least provide code for a minimal reproduction of the main insights
% of your paper).
%
% Please do not modify the questions and only use the provided macros for your
% answers. Note that this section does not count towards the page limit.

\begin{enumerate}
\item For all authors\dots
  \begin{enumerate}
  \item Do the main claims made in the abstract and introduction accurately
reflect the paper's contributions and scope?
\answerYes{Our claims have been checked against the evidence we provide.}
  \item Did you describe the limitations of your work?
\answerYes{Yes, we discuss limitations in \cref{sec:limitations}.}
  \item Did you discuss any potential negative societal impacts of your work?
\answerYes{Yes, we discuss impacts in \cref{sec:impact}.}
  \item Have you read the ethics author's and review guidelines and ensured that
your paper conforms to them? \url{https://automl.cc/ethics-accessibility/}
\answerYes{We have read the guidelines and done our best to follow them.}
  \end{enumerate}
\item If you are including theoretical results\dots
  \begin{enumerate}
  \item Did you state the full set of assumptions of all theoretical results?
\answerNA{Not applicable.}
  \item Did you include complete proofs of all theoretical results?
\answerNA{Not applicable.}
  \end{enumerate}
\item If you ran experiments\dots
  \begin{enumerate}
  \item Did you include the code, data, and instructions needed to reproduce the
main experimental results, including all requirements (e.g.,
\texttt{requirements.txt} with explicit version), an instructive
\texttt{README} with installation, and execution commands (either in the
supplemental material or as a \textsc{url})?
\answerYes{The anonymized repository can be found at \url{https://anon-github.automl.cc/r/better-da-4936}}
  \item Did you include the raw results of running the given instructions on the
given code and data?
\answerYes{We have provided guidance on what the results should look like when running the code.}
  \item Did you include scripts and commands that can be used to generate the
figures and tables in your paper based on the raw results of the code, data,
and instructions given?
\answerYes{Full instructions for each step of the process have been included.}
  \item Did you ensure sufficient code quality such that your code can be safely
executed and the code is properly documented?
\answerYes{We have tried to ensure good code quality and documentation.}
  \item Did you specify all the training details (e.g., data splits,
pre-processing, search spaces, fixed hyperparameter settings, and how they
were chosen)?
\answerYes{We have been careful to specify all training details, see especially \cref{sec:training_details}.}
  \item Did you ensure that you compared different methods (including your own)
exactly on the same benchmarks, including the same datasets, search space,
code for training and hyperparameters for that code?
\answerYes{One main contribution of this work is to perform fair benchmarking of existing literature and guide future work towards better practices in this regard.}
  \item Did you run ablation studies to assess the impact of different
components of your approach?
\answerYes{In the sense that this paper is focused on benchmarking, we have looked at different perspectives of the process, e.g.~looking at validators, data splits etc.}
  \item Did you use the same evaluation protocol for the methods being compared?
\answerYes{This is a main contribution of this work.}
  \item Did you compare performance over time?
\answerNo{All methods we compare in this work (within a given setting) have similar running times so that aspect of the training procedure was not considered important to explore further.}
  \item Did you perform multiple runs of your experiments and report random seeds?
\answerNo{But we instead perform our benchmark over many datasets, domains and hyperparameter choices.}
  \item Did you report error bars (e.g., with respect to the random seed after
running experiments multiple times)?
\answerYes{Error bars with respect to algorithms and datasets have been included in \cref{sec:correlations}.}
  \item Did you use tabular or surrogate benchmarks for in-depth evaluations?
\answerNo{The focus of our benchmarking study is on more realistic datasets, although simpler ones like MNIST-M, MNIST-MR are included.}
  \item Did you include the total amount of compute and the type of resources
used (e.g., type of \textsc{gpu}s, internal cluster, or cloud provider)?
\answerYes{We have estimated the amount of compute necessary for our experiments in \cref{sec:compute}.}
  \item Did you report how you tuned hyperparameters, and what time and
resources this required (if they were not automatically tuned by your AutoML
method, e.g. in a \textsc{nas} approach; and also hyperparameters of your
own method)?
\answerYes{This is included in the estimation in \cref{sec:compute}.}
  \end{enumerate}
\item If you are using existing assets (e.g., code, data, models) or
  curating/releasing new assets\dots
  \begin{enumerate}
  \item If your work uses existing assets, did you cite the creators?
\answerYes{All assets have been cited.}
  \item Did you mention the license of the assets?
\answerYes{The licenses of the assets we use are listed in \ref{sec:assets}.}
  \item Did you include any new assets either in the supplemental material or as
a \textsc{url}?
\answerYes{The MNIST-MR dataset can be generated using the code provided.}
  \item Did you discuss whether and how consent was obtained from people whose
data you're using/curating?
\answerNA{No people's personal data was used.}
  \item Did you discuss whether the data you are using/curating contains
personally identifiable information or offensive content?
\answerNA{No people's personal data was used.}
  \end{enumerate}
\item If you used crowdsourcing or conducted research with human subjects\dots
  \begin{enumerate}
  \item Did you include the full text of instructions given to participants and
screenshots, if applicable?
\answerNA{Not applicable.}
  \item Did you describe any potential participant risks, with links to
Institutional Review Board (\textsc{irb}) approvals, if applicable?
\answerNA{Not applicable.}
  \item Did you include the estimated hourly wage paid to participants and the
total amount spent on participant compensation?
\answerNA{Not applicable.}
  \end{enumerate}
\end{enumerate}
}

% content will be automatically hidden during submission
\begin{acknowledgements}

\end{acknowledgements}

% print bibliography -- for bibtex / natbib, use:

% \bibliography{...}

% and for biber / biblatex, use:

% \printbibliography

% supplemental material -- everything hereafter will be suppressed during
% submission time if the hidesupplement option is provided!

\newpage

\appendix

\section{Regression} \label{sec:regression}
\keypoint{Setup}
We construct a regression dataset with a domain shift akin to MNIST-M \cite{Glorot2010UnderstandingNetworks}. The source domain consists of 32x32 images where, for each image, a single digit taken from MNIST is pasted onto a black background. The digit is randomly scaled between 4x4 and 16x16 pixels and its location is randomised while ensuring the entire digit is visible. The label accompanying the image is the top left $(x_1, y_1)$ and bottom right $(x_2, y_2)$ coordinates of the digit bounding box. The target domain is constructed similarly, but instead of a black background, we use 32x32 regions cropped from the BSDS500 dataset \cite{bsds500}.

We discretize the label space as follows. The bounding box labels and predictions take the following form \{$x_1, y_1, x_2, y_2$\}, where each element is a real value between 0 and 1. The function $q$ discretizes each element in the vector into one of 8 uniformly spaced classes between 0 and 1. The class of the full vector is then $c = q(x_1) + 8 q(y_1) + 8^2 q(x_2) + 8^3 q(y_2)$. This transformation into class values is performed for all labels and predictions.

The architecture is the same as used for the MNIST-M experiments in the classification setting, with an adjusted final layer for regression. The loss used on the source data is the mean squared error. All other details are the same as in the above UDA setting.
% \keypoint{Algorithms}
For this setup we train six algorithms, ADDA \cite{adda}, CORAL \cite{coral}, DANN \cite{ganin2016domainAdversarialNN}, GAN \cite{Goodfellow2014GenerativeNets}, MMD \cite{mmd}, VADA \cite{vada}.
% \keypoint{Validators}
Many of the validators we have considered so far rely on categorical labels and predictions. In this regression setup we, therefore, discretize the label space as described above.

\begin{table*}[t]
\centering
\caption{Comparison of validation criteria for model selection in UDA for a regression task. We report the target test MSE for the top models selected by each validator. Lower is better. The colour of a cell indicates whether that model/validator combination beats the source-only model (green) or not (red), with a darker red colour meaning it gets more than twice the MSE of the source-only model performance.}
\label{tab:regression}
\resizebox{1.0\linewidth}{!}{
\begin{tabular}{l|ccccccccccccccc|c}
\toprule
{}                   & RankMe                        & AMI                           & ARI                           & V-Measure                     & FMI                           & Silhouette                    & DBI                           & CHI                                                   & BNM                                                   & MMD                           & CORAL                         & SND                           & IM                                                    & Entropy                       & MSE                           & Oracle                        \\ \hline
ADDA                   & \cellcolor[HTML]{E9FFC0}17.52 & \cellcolor[HTML]{E9FFC0}43.96 & \cellcolor[HTML]{E9FFC0}44.44 & \cellcolor[HTML]{E9FFC0}43.96 & \cellcolor[HTML]{E9FFC0}44.44 & \cellcolor[HTML]{E9FFC0}46.69 & \cellcolor[HTML]{E9FFC0}16.50 & \cellcolor[HTML]{E9FFC0}18.31                         & \cellcolor[HTML]{E9FFC0}46.69                         & \cellcolor[HTML]{E9FFC0}16.76 & \cellcolor[HTML]{E9FFC0}17.49 & \cellcolor[HTML]{E9FFC0}17.52 & \cellcolor[HTML]{E9FFC0}17.49                         & \cellcolor[HTML]{E9FFC0}16.50 & \cellcolor[HTML]{E9FFC0}46.27 & \cellcolor[HTML]{E9FFC0}16.50 \\
CORAL                  & \cellcolor[HTML]{E9FFC0}45.06 & \cellcolor[HTML]{E9FFC0}45.12 & \cellcolor[HTML]{E9FFC0}46.09 & \cellcolor[HTML]{E9FFC0}45.12 & \cellcolor[HTML]{E9FFC0}46.09 & \cellcolor[HTML]{E9FFC0}45.22 & \cellcolor[HTML]{E9FFC0}44.98 & \cellcolor[HTML]{E9FFC0}46.68                         & \cellcolor[HTML]{FFD1D1}47.02                         & \cellcolor[HTML]{FFD1D1}47.02 & \cellcolor[HTML]{FFD1D1}47.02 & \cellcolor[HTML]{E9FFC0}45.71 & \cellcolor[HTML]{FFD1D1}47.02                         & \cellcolor[HTML]{FFD1D1}47.02 & \cellcolor[HTML]{FFD1D1}47.11 & \cellcolor[HTML]{E9FFC0}43.60 \\
DANN                   & \cellcolor[HTML]{FFD1D1}81.63 & \cellcolor[HTML]{E9FFC0}42.17 & \cellcolor[HTML]{E9FFC0}42.17 & \cellcolor[HTML]{E9FFC0}42.17 & \cellcolor[HTML]{E9FFC0}46.95 & \cellcolor[HTML]{FFD1D1}79.15 & \cellcolor[HTML]{FFD1D1}54.40 & \cellcolor[HTML]{FFD1D1}55.18                         & \cellcolor[HTML]{FFD1D1}74.76                         & \cellcolor[HTML]{FFD1D1}81.63 & \cellcolor[HTML]{FFD1D1}53.99 & \cellcolor[HTML]{E9FFC0}35.22 & \cellcolor[HTML]{FFD1D1}81.63                         & \cellcolor[HTML]{FFD1D1}54.32 & \cellcolor[HTML]{E9FFC0}45.41 & \cellcolor[HTML]{E9FFC0}35.22 \\
GAN                    & \cellcolor[HTML]{E9FFC0}16.29 & \cellcolor[HTML]{E9FFC0}42.06 & \cellcolor[HTML]{E9FFC0}43.37 & \cellcolor[HTML]{E9FFC0}42.06 & \cellcolor[HTML]{E9FFC0}43.22 & \cellcolor[HTML]{FFD1D1}53.75 & \cellcolor[HTML]{FFD1D1}47.28 & \cellcolor[HTML]{FFD1D1}49.41                         & \cellcolor[HTML]{FFD1D1}47.60                         & \cellcolor[HTML]{E9FFC0}16.29 & \cellcolor[HTML]{E9FFC0}16.80 & \cellcolor[HTML]{FFD1D1}53.74 & \cellcolor[HTML]{E9FFC0}40.53                         & \cellcolor[HTML]{E9FFC0}31.59 & \cellcolor[HTML]{E9FFC0}44.10 & \cellcolor[HTML]{E9FFC0}16.29 \\
MMD                    & \cellcolor[HTML]{FFD1D1}64.96 & \cellcolor[HTML]{E9FFC0}46.86 & \cellcolor[HTML]{E9FFC0}46.94 & \cellcolor[HTML]{E9FFC0}46.86 & \cellcolor[HTML]{E9FFC0}46.94 & \cellcolor[HTML]{FFD1D1}90.67 & \cellcolor[HTML]{FFD1D1}55.63 & \cellcolor[HTML]{FFA6A6}{\color[HTML]{333333} 138.98} & \cellcolor[HTML]{FFA6A6}{\color[HTML]{333333} 138.98} & \cellcolor[HTML]{FFD1D1}64.96 & \cellcolor[HTML]{FFD1D1}63.46 & \cellcolor[HTML]{E9FFC0}45.41 & \cellcolor[HTML]{FFA6A6}{\color[HTML]{333333} 138.98} & \cellcolor[HTML]{FFD1D1}90.67 & \cellcolor[HTML]{FFD1D1}61.47 & \cellcolor[HTML]{E9FFC0}45.12 \\
VADA                   & \cellcolor[HTML]{E9FFC0}17.54 & \cellcolor[HTML]{E9FFC0}44.69 & \cellcolor[HTML]{FFD1D1}57.33 & \cellcolor[HTML]{E9FFC0}44.69 & \cellcolor[HTML]{FFD1D1}57.33 & \cellcolor[HTML]{FFD1D1}48.26 & \cellcolor[HTML]{E9FFC0}17.54 & \cellcolor[HTML]{E9FFC0}41.73                         & \cellcolor[HTML]{FFD1D1}51.65                         & \cellcolor[HTML]{E9FFC0}14.38 & \cellcolor[HTML]{E9FFC0}14.95 & \cellcolor[HTML]{FFD1D1}46.27 & \cellcolor[HTML]{E9FFC0}14.38                         & \cellcolor[HTML]{FFD1D1}55.82 & \cellcolor[HTML]{FFD1D1}48.44 & \cellcolor[HTML]{E9FFC0}14.38 \\
\midrule
Avg. $\downarrow$      & \cellcolor[HTML]{E9FFC0}40.50 & \cellcolor[HTML]{E9FFC0}44.14 & \cellcolor[HTML]{E9FFC0}46.72 & \cellcolor[HTML]{E9FFC0}44.14 & \cellcolor[HTML]{FFD1D1}47.50 & \cellcolor[HTML]{FFD1D1}60.62 & \cellcolor[HTML]{E9FFC0}39.39 & \cellcolor[HTML]{FFD1D1}58.38                         & \cellcolor[HTML]{FFD1D1}67.78                         & \cellcolor[HTML]{E9FFC0}40.17 & \cellcolor[HTML]{E9FFC0}35.62 & \cellcolor[HTML]{E9FFC0}40.64 & \cellcolor[HTML]{FFD1D1}56.67                         & \cellcolor[HTML]{FFD1D1}49.32 & \cellcolor[HTML]{FFD1D1}48.80 & \cellcolor[HTML]{E9FFC0}28.52 \\
Avg. Rank $\downarrow$ & 6.33                          & 5.42                          & 8.42                          & 5.42                          & 8.58                          & 11.33                         & 5.50                          & 10.00                                                 & 12.58                                                 & 6.92                          & 6.25                          & 6.25                          & 8.50                                                  & 8.33                          & 10.17                         & -                             \\
Correlation $\uparrow$ & -0.31                         & 0.13                          & -0.09                         & 0.13                          & -0.16                         & -0.17                         & 0.05                          & 0.12                                                  & -0.25                                                 & 0.17                          & 0.12                          & -0.03                         & -0.07                                                 & -0.25                         & -0.29                         & -                             \\ \hline
Source-only            & \cellcolor[HTML]{FFD1D1}49.06 & \cellcolor[HTML]{FFD1D1}49.06 & \cellcolor[HTML]{FFD1D1}49.06 & \cellcolor[HTML]{FFD1D1}49.06 & \cellcolor[HTML]{FFD1D1}49.06 & \cellcolor[HTML]{FFD1D1}49.06 & \cellcolor[HTML]{FFD1D1}49.06 & \cellcolor[HTML]{FFD1D1}49.06                         & \cellcolor[HTML]{FFD1D1}49.06                         & \cellcolor[HTML]{FFD1D1}49.06 & \cellcolor[HTML]{FFD1D1}49.06 & \cellcolor[HTML]{FFD1D1}49.06 & \cellcolor[HTML]{FFD1D1}49.06                         & \cellcolor[HTML]{FFD1D1}49.06 & \cellcolor[HTML]{C0C0C0}46.96 & \cellcolor[HTML]{E9FFC0}42.28 \\
\bottomrule
\end{tabular}
}
\end{table*}

\keypoint{Results} Table~\ref{tab:regression} shows the results on this regression task.  

\ques{Are conclusions still valid beyond image classification?}
The overall results show a similar trend to the observation of UDA for image classification. 1) Now, CORAL works as the best validator. However, there is no one validator working consistently well for all methods. 2) CORAL, as a UDA method, works most robustly with all validation criteria, leading to all selected results close to its oracle performance, which, though, is not ideal. However, we can see now the correlations are very low for all validators, indicating that there is no reliable validator in this case that works robustly to select a good UDA model.

\section{Assets} \label{sec:assets}
\keypoint{Code} Our anonymized code base is available at \url{https://anon-github.automl.cc/r/better-da-4936}. In this work, we make use of the \texttt{KevinMusgrave/pytorch-adapt}, \texttt{DequanWang/tent}, \texttt{vita-epfl/ttt-plus-plus} and \texttt{matthijsz/weightedcorr} libraries, all available on GitHub and all released under the MIT License.

\keypoint{Data} The creators of the MNIST \cite{LeCun2010MNISTDatabase}, MNIST-M \cite{ganin2016domainAdversarialNN} and Office-31 \cite{Saenko2010AdaptingDomains} datasets have not provided obvious licenses, but both datasets were created for open academic use. Both VisDA-2017 \cite{visda2017} and Office-Home \cite{officehome} are released under custom licenses allowing non-commercial research and use for educational purposes.

\section{Training Details} \label{sec:training_details}
\subsection{Settings}
To fully clarify our adaptation settings, we present in algorithms \ref{alg:uda}, \ref{alg:sfda} and \ref{alg:tta} the benchmarking procedure for UDA, SFDA and TTA, respectively.

\begin{algorithm}
\caption{UDA benchmarking setup.}\label{alg:uda}
\begin{algorithmic}
\Require Source data $\gD_S$, target data $\gD_T$, algorithm $\gA$, parameters $\vtheta$, hyperparameter search space $\sH$, validator $V$.
\For {hyperparameters $\vh \sim \sH$} \Comment{Sample hyperparameters}
    \State $\vtheta^*_{\vh} = \argmin_{\vtheta} \gA(\vtheta, \gD_S, \gD_T; \vh)$ \Comment{Optimise model}
\EndFor
\State $\vh^* = \argmax_{\vh} V(\vtheta^*_{\vh}, \gD_S, \gD_T)$ \Comment{Select best hyperparameters}
\end{algorithmic}
\end{algorithm}

\begin{algorithm}
\caption{SFDA benchmarking setup.}\label{alg:sfda}
\begin{algorithmic}
\Require Target data $\gD_T$, algorithm $\gA$, parameters $\vtheta$, hyperparameter search space $\sH$, validator $V$.
\For {hyperparameters $\vh \sim \sH$} \Comment{Sample hyperparameters}
    \State $\vtheta^*_{\vh} = \argmin_{\vtheta} \gA(\vtheta, \gD_T; \vh)$ \Comment{Optimise model}
\EndFor
\State $\vh^* = \argmax_{\vh} V(\vtheta^*_{\vh}, \gD_T)$ \Comment{Select best hyperparameters}
\end{algorithmic}
\end{algorithm}

\begin{algorithm}
\caption{TTA benchmarking setup.}\label{alg:tta}
\begin{algorithmic}
\Require Test data $\gD_T$, algorithm $\gA$, parameters $\vphi$, hyperparameter search space $\sH$, validator $V$.
\For {each batch $X \sim \gD_T$}
    \For {hyperparameters $\vh \sim \sH$} \Comment{Sample hyperparameters}
\State $\vtheta = \vphi$ \Comment{Reset model}
\State $\vtheta^*_{\vh} = \argmin_{\vtheta} \gA(\vtheta, X; \vh)$ \Comment{Optimise model}
    \EndFor
    \State $\vh^*_X = \argmax_{\vh} V(\vtheta^*_{\vh}, X)$ \Comment{Select best hyperparameters}
\EndFor
\end{algorithmic}
\end{algorithm}

\keypoint{Data splits}
We split all domains of all datasets into train (60\%), val (20\%) and test (20\%) splits.

\keypoint{Optimisation}
The optimiser for UDA classification, UDA regression, and SFDA is Adam with parameters \texttt{\{betas=(0.9, 0.999)\}} and weight decay of $1e-4$. For TTA the optimizer is SGD with a momentum of $0.9$ (the optimizer is reset after each batch, like the model parameters). The learning rate across all settings is sampled from a log-uniform distribution over \texttt{[1e-5, 1e-1]}.

We train for 100 epochs for MNIST-M and MNIST-MR, 200 on VisDA-2017 and Office-Home. On Office-31 it is 200 if amazon is the target and 2000 otherwise. The number of saved checkpoints is always 20. For our episodic TTA setup, we perform 20 updates on each batch, saving a checkpoint after each update.

\keypoint{Architecture}
The backbone for experiments on MNIST-M and our regression version MNIST-MR is a LeNet-5\footnote{The LeNet backbone consists of the convolutional block \{Conv, ReLU, MaxPool, Conv, ReLU, MaxPool\}.} \cite{lenet}, and for all other experiments, a ResNet50 \cite{He2016DeepRecognition}. The classifier/regressor is an MLP with two blocks of \{Linear, ReLU, Dropout\} followed by a final linear layer.

\keypoint{Source Training}
The source model consists of the backbone and classifier/regressor as defined above. When using a ResNet50 backbone, we initialise it with ImageNet pre-trained weights (available in PyTorch \cite{Paszke2017AutomaticPyTorch} as \texttt{resnet50(weights=ResNet50\_Weights.IMAGENET1K\_V1)} and freeze the backbone during source training, thereby only updating the classification head. When using a LeNet backbone, we update the entire network during source training. For TTA on CIFAR10-C we use the pre-trained CIFAR10 checkpoint provided by \cite{ttt_plus_plus} as initialisation. In all other cases, the best model checkpoint as selected by source validation accuracy is used as the initialisation for all adaptation algorithms.

\keypoint{Adaptation}
During UDA adaptation, the model receives a batch consisting of 64 source examples (with labels) and 64 target examples (without labels). For SFDA, the model only receives the target examples.

\subsection{Hyperparameters}
Throughout the experiments conducted in this work, we perform random search for finding the best hyperparameters, with 10 random choices per algorithm.
\rebuttal{Better performance can potentially be reached by using e.g.~BayesOpt. In this work, we focus on analysis and prefer the simpler random search to (1) enable computing a correlation score between the validation criteria and test performance (correlation computed over both high and low-quality checkpoints), as we report in our main tables. Also (2) because our comparisons involve comparing the “best” possible checkpoint with the one discovered by each validator, we did not want to risk aggressively optimising a bad validator and thus having no good checkpoints available for selection by the oracle.} The hyperparameter search spaces for all algorithms are specified in \cref{tab:hyperparameters}. Whenever possible, these are identical to those used in \cite{Musgrave2020UnsupervisedCheck}.

\begin{table}[p]
\centering
\caption{Hyperparameter search spaces for all algorithms considered. Some algorithms are used in multiple settings (e.g. DANN and MMD are used for UDA classification and regression and SHOT is used for SFDA and TTA). In such cases, the search spaces are the same across settings.}
\label{tab:hyperparameters}
\begin{tabular}{ccc}
\toprule
Algorithm & Hyperparameter & Search Space \\
\midrule
\multirow{3}{*}{ATDOC} & $\lambda_{atdoc}$ & \texttt{[0, 1]} \\
 & $K_{atdoc}$   & \texttt{int([5, 25], step=5)} \\
 & $\lambda_{L}$ & \texttt{[0, 1]} \\
\midrule
\multirow{2}{*}{BNM}   & $\lambda_{bnm}$   & \texttt{[0, 1]} \\
 & $\lambda_{L}$ & \texttt{[0, 1]} \\
\midrule
\multirow{3}{*}{DANN}  & $\lambda_{D}$ & \texttt{[0, 1]} \\
 & $\lambda_{grl}$   & \texttt{log([0.1, 10])} \\
 & $\lambda_{L}$ & \texttt{[0, 1]} \\
\midrule
\multirow{3}{*}{MCC}   & $\lambda_{mcc}$   & \texttt{[0, 1]} \\
 & $T_{mcc}$ & \texttt{[0.2, 5])} \\
 & $\lambda_{L}$ & \texttt{[0, 1]} \\
\midrule
\multirow{3}{*}{MCD}   & $N_{mcd}$ & \texttt{int([1, 10])} \\
 & $\lambda_{disc}$  & \texttt{[0, 1]} \\
 & $\lambda_{L}$ & \texttt{[0, 1]} \\
\midrule
\multirow{3}{*}{MMD}   & $\lambda_{F}$ & \texttt{[0, 1]} \\
 & $\gamma_{exp}$& \texttt{int([1, 8])} \\
 & $\lambda_{L}$ & \texttt{[0, 1]} \\
\midrule
\midrule
\multirow{2}{*}{ADDA}  & $\lambda_{D}$ & \texttt{[0, 1]} \\
 & $\lambda_{G}$ & \texttt{[0, 1]} \\
\midrule
\multirow{2}{*}{CORAL} & $\lambda_{F}$ & \texttt{[0, 1]} \\
 & $\lambda_{L}$ & \texttt{[0, 1]} \\
\midrule
\multirow{3}{*}{GAN}   & $\lambda_{D}$ & \texttt{[0, 1]} \\
 & $\lambda_{G}$ & \texttt{[0, 1]} \\
 & $\lambda_{L}$ & \texttt{[0, 1]} \\
\midrule
\multirow{5}{*}{VADA}  & $\lambda_{D}$ & \texttt{[0, 1]} \\
 & $\lambda_{G}$ & \texttt{[0, 1]} \\
 & $\lambda_{V}$ & \texttt{[0, 1]} \\
 & $\lambda_{E}$ & \texttt{[0, 1]} \\
 & $\lambda_{L}$ & \texttt{[0, 1]} \\
\midrule
\midrule
\multirow{2}{*}{AAD}   & $\lambda_{aad}$ & \texttt{[0, 1]} \\
 & $K_{aad}$   & \texttt{int([3, 5]} \\
\midrule
\multirow{3}{*}{NRC}   & $\lambda_{L}$ & \texttt{[0, 1]} \\
 & $K_{nrc}$   & \texttt{int([2, 5]} \\
 & $KK_{nrc}$   & \texttt{int([2, 5]} \\
\midrule
\multirow{3}{*}{SHOT}  & $\lambda_{cls}$   & \texttt{[0, 1]} \\
 & $\lambda_{ent}$   & \texttt{[0, 1]} \\
 & $\lambda_{L}$ & \texttt{[0, 1]} \\
\midrule
\midrule
\multirow{1}{*}{TENT}  & $\lambda_{L}$ & \texttt{[0, 1]} \\
\midrule
\bottomrule
\end{tabular}
\end{table}

\section{Validation Details} \label{sec:validators}
\subsection{Validators}
A recent work systematically investigated the possible validation criteria for UDA, which we summarise below using $\hat{\vy}$ to denote the one-hot predictions of the model and $\vy$ as the one-hot ground truth labels.
\keypoint{Source accuracy} $d$ is simply the accuracy metric and $\gD_V$ can be a training or validation set from a source domain. 
  \begin{equation}
 d(f_\vtheta, \gD_V) = \frac{1}{N_V} \sum_{i=1}^{N_V} \vone(\hat{\vy} = \vy),
  \end{equation}
where $\vone(\cdot)$ is the indicator function that evaluates to one if its argument is true and zero otherwise.
\keypoint{Entropy} Entropy has been used in an adaptation loss~\cite{wang2021tent} as well as for model selection. In this case, $d$ computes the confidence of the model predictions, as measured by the entropy of the predicted label distribution, and $\gD_V$ is typically the training or validation set from an unlabelled target domain. We further investigate the effect when $\gD_V$ comes from the source domain.
  \begin{equation}
  \begin{aligned}
 d(f_\vtheta, \gD_V) = \frac{1}{N_V} \sum_{i=1}^{N_V} H(\vp_i), \vp_i = f_\vtheta(\vx_i),
  \end{aligned}
  \end{equation}
where
  \begin{equation}
 H(\vp) = -\sum_{j=1}^K \evp_{[j]} \textup{log} \evp_{[j]},
  \end{equation}
computes the entropy of the categorical distribution, $\vp$.
\keypoint{Information maximisation (IM)} IM is often used as an adaptation loss as well~\cite{Shi2012Information-TheoreticalAdaptation} to maximise the diversity of prediction in addition to confidence.
  \begin{equation}
 d(f_\vtheta, \gD_V) = H(\frac{1}{N_V} \sum_{i=1}^{N_V} \vp_i) - \frac{1}{N_v} \sum_{i=1}^{N_v} H(\vp_i).
  \end{equation}
\keypoint{Adjusted Mutual Information (AMI)} This is the adjusted mutual information between predicted and cluster labels. 
\begin{equation}
 d(f_\vtheta, \gD_V) = \operatorname{AMI}(\vp, \operatorname{CL}(\gD_V))
  \end{equation}
where $\operatorname{CL}(\gD_V)$ is the cluster labels for validation set $\gD_V$, which can be the target training or validation set.

\keypoint{V-Measure} Similarly to AMI, this is a metric defined over clustering labels and predictions. It is defined as the harmonic mean between homogeneity and completeness \cite{vmeasure}.

\keypoint{Other clustering measures} Along with AMI and V-Measure, we compute several other related clustering measures, namely, adjusted Rand index, Fowlkes–Mallows index, silhouette score, Davies–Bouldin index and Calinski-Harabasz index.

\keypoint{RankMe} Originally proposed for estimating the transferability of self-supervised representations \cite{rankme}, RankMe approximates the rank of the feature matrix on pre-training data. We investigate its application to both source and target domain data.

\keypoint{CORAL} CORAL is an adaptation algorithm that aligns the feature distributions of the source and target data by minimising second-order statistics \cite{coral}. Their loss can be used as a validator and is defined as the difference between the covariance matrices of the two domains, $C_S$ and $C_T$.
\begin{equation}
%\begin{aligned}
d(f_\vtheta, \gD_V) = \operatorname{CORAL}(\gD_S, \gD_T) = \frac{1}{4 d^2} \|C_S - C_T\|^2_F \\
%\end{aligned}
\end{equation}

\keypoint{Maximum mean discrepancy (MMD)} A common metric used to compute the discrepancy of feature distributions from source and target domains~\cite{mmd}, which can be used with the assumption that the trained model may have a good target performance when the source and target domain features are aligned.
\begin{equation}
\begin{aligned}
  & d(f_\vtheta, \gD_V) = \operatorname{MMD}(\gD_S, \gD_T) \\
  & = \frac{1}{N_S(N_S-1)}\sum_{i=1}^{N_S}\sum_{j\neq i}^{N_S}k(\vs_i, \vs_j; f_\vtheta) \\
  &+ \frac{1}{N_T(N_T-1)}\sum_{i=1}^{N_T}\sum_{j\neq i}^{N_T}k(\vt_i, \vt_j) \\
  & - \frac{2}{N_SN_T}\sum_{i=1}^{N_S}\sum_{j= 1}^{N_T}k(\vs_i, \vt_j), \\
  & k(a, b) = \exp{\frac{-\| a - b \|^2_2}{e}},
\end{aligned}
\end{equation}
where $\vs$ and $\vt$ are the features extracted for the data from source and target domains, respectively.
\cut{
And additionally, the \tb{Class-wise MMD} which computes the MMD distance between source and target domains from the same classes separately.
\begin{equation}
\begin{aligned}
  d(f_\vtheta, \gD_V) & = \operatorname{CW-MMD}(\gD_S, \gD_T)
  \\ 
  &= \frac{1}{C} \sum_{i=1}^C \operatorname{MMD}(\gD_S|_{\vy=i}, \gD_T|_{\vy=i})
\end{aligned}
\end{equation}
}
When MMD is used for validation, the validation set combines the train sets or validation sets of source and target domains.
\keypoint{Soft neighbourhood density (SND)} SND computes the entropy based on the gram matrix of the validation features.
\begin{equation}
\begin{aligned}
  & d(f_\vtheta, \gD_V) = H(\alpha(\mX, \tau)), \\
  & \mX = \bm{v}^{T} \bm{v},
\end{aligned}
\end{equation}
where $\bm{v}$ are the data features, $\alpha(\cdot)$ and $\tau$ are softmax function and temperature. Here $\gD_V$ can be the train or validation set of source or target domains.
  % \item \tb{Rev}, which generates the pseudo labels for the target domain using the training UDA model and reverses the process and treats the source domain as the virtual target domain while the pseudo-labelled target as the virtual source. Then, the model can be validated by this alternating training process.
\keypoint{Batch nuclear-norm maximisation (BNM)} BNM was originally a UDA algorithm, which maximises the nuclear norm of the prediction matrix in a batch, being repurposed as a validation criterion.
\begin{equation}
 d(f_\vtheta, \gD_V) = \| \mP \|_{*},
 \mP = f_\vtheta(\gD_V)
  \end{equation}
where $\mP \in R^{N_V\times C}$ the prediction matrix of whole data in $\gD_V$ using $f_\vtheta$. And $\|\|_*$ computes the nuclear norm.

\subsection{\rebuttal{Time Complexity}}
\begin{figure}
    \centering
    \includegraphics{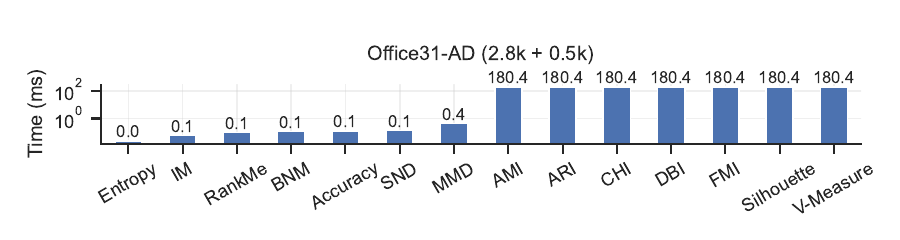}
    \includegraphics{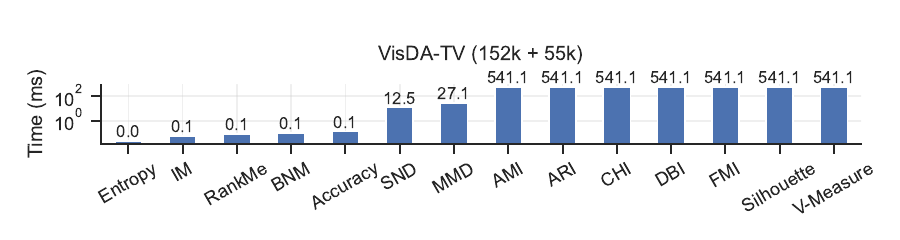}
    \caption{\rebuttal{Computation time of UDA validators on (top) Office31-AD and (bottom) VisDA-TV. The clustering-based validators are all significantly more compute-intensive, though it still takes only half a second to compute for a total of 200k datapoints on the VisDA dataset.}}
    \label{fig:time_complexity}
\end{figure}

\rebuttal{Most validators are very quick to compute, requiring only a loop through the features, logits or predictions or some matrix multiplications on the same. Those requiring the extra clustering step (AMI, ARI, CHI, DBI, FMI, V-Measure and Silhouette) all take significantly longer. Nonetheless, no validator is prohibitively expensive compared to the time required to adapt the models. See \cref{fig:time_complexity} for numbers on two representative datasets.}

\subsection{Validator Versions in Main Paper}
For the tables and figures in the main document, we present a single version of each validator, the one that gives the highest performance when averaged over algorithms and datasets. However, there are multiple options for each, for example, which data split is used or whether we use features, logits or prediction vectors to compute the score. \Cref{tab:validator_versions} shows which versions are used for each validator.

\begin{sidewaystable}[p]
\centering
\caption{Summary of which version of each validator is presented in the tables of the main document. $\mS_V$: source val split, $\mT_T$: target train split, $\mT_V$: target val split.}
\label{tab:validator_versions}
\resizebox{1.0\linewidth}{!}{
\begin{tabular}{l|l|ccccccccccccccc}
\toprule
Setting & Option & RankMe &  AMI &  ARI &  V-Measure &  FMI &  Silhouette &  DBI &  CHI &  BNM &  MMD & CORAL &  -SND &  IM &  Entropy &  Accuracy/MSE \\
\midrule
\multirow{2}{*}{UDA (Classification)} & Split & $\mS_V+\mT_V$ & $\mS_V+\mT_V$ & $\mS_V+\mT_V$ & $\mS_V+\mT_V$ & $\mS_V+\mT_V$ & $\mS_V+\mT_T$ & $\mS_V+\mT_V$ & $\mS_V+\mT_V$ & $\mS_V+\mT_T$ & $\mS_V+\mT_V$ & $\mS_V+\mT_V$ & $\mT_T$ & $\mS_V+\mT_T$ & $\mS_V+\mT_T$ & $\mS_V$ \\
 & Layer & Predictions & Logits & Logits & Logits & Logits & Logits & Logits & Logits & Predictions & Predictions & Predictions & Predictions & Predictions & Predictions & Predictions \\
\midrule
\multirow{2}{*}{UDA (Regression)}& Split & $\mT_V$ & $\mS_V+\mT_T$ & $\mS_V+\mT_T$ & $\mS_V+\mT_T$ & $\mS_V+\mT_V$ & $\mS_V+\mT_T$ & $\mS_V+\mT_V$ & $\mS_V+\mT_T$ & $\mT_T$ & $\mS_V+\mT_V$ & $\mS_V+\mT_T$ & $\mT_V$ & $\mT_T$ & $\mT_T$ & $\mS_V$ \\
 & Layer & Predictions & Features & Features & Features & Features & Features & Features & Logits & Predictions & Predictions & Features & Predictions & Predictions & Predictions & Predictions \\
\midrule
\multirow{2}{*}{SFDA}    & Split & $\mT_T$ & $\mT_T$ & $\mT_V$ & $\mT_T$ & $\mT_T$ & $\mT_T$ & $\mT_T$ & $\mT_T$ & $\mT_T$ & - & - & $\mT_V$ & $\mT_T$ & $\mT_T$ & - \\
 & Layer & Predictions & Features & Logits & Features & Features & Logits & Logits & Features & Predictions & - & - & Predictions & Predictions & Predictions & - \\
\midrule
\multirow{2}{*}{TTA (CIFAR10-C)} & Split & $\mT_T$ & $\mT_T$ & $\mT_T$ & $\mT_T$ & $\mT_T$ & $\mT_T$ & $\mT_T$ & $\mT_T$ & $\mT_T$ & - & - & $\mT_T$ & $\mT_T$ & $\mT_T$ & - \\
 & Layer & Predictions & Logits & Logits & Logits & Logits & Logits & Features & Logits & Predictions & - & - & Predictions & Predictions & Predictions & - \\
\midrule
\multirow{2}{*}{TTA (Office-Home)}    & Split & $\mT_T$ & $\mT_T$ & $\mT_T$ & $\mT_T$ & $\mT_T$ & $\mT_T$ & $\mT_T$ & $\mT_T$ & $\mT_T$ & - & - & $\mT_T$ & $\mT_T$ & $\mT_T$ & - \\
 & Layer & Logits & Features & Features & Features & Features & Features & Features & Features & Predictions & - & - & Predictions & Predictions & Predictions & - \\
\bottomrule
\end{tabular}
}
\end{sidewaystable}
% SFDA
% target_train_preds_rank_me_score - RankMe (Target Train Preds): 
% target_train_class_ami_score - AMI (Target Train Features): 
% target_val_logits_class_ari_score - ARI (Target Val Logits): 
% target_train_class_v_measure_score - V-Measure (Target Train Features): 
% target_train_class_fmi_score - FMI (Target Train Features): 
% target_train_logits_class_silhouette_score - Silhouette (Target Train Logits): 
% target_train_logits_class_dbi_score - BDI (Target Train Logits): 
% target_train_class_chi_score - CHI (Target Train Features): 
% target_train_bnm_score - BNM (Target Train): 
% target_val_neg_snd_score - -SND (Target Val): 
% target_train_im_score - IM (Target Train): 
% target_train_entropy_score - Entropy (Target Train): 

% TTA-Office-Home
% validator versions for ++ adaptors:
%['target_train_logits_rank_me_score',
% 'target_train_class_ami_score',
% 'target_train_class_ari_score',
% 'target_train_class_v_measure_score',
% 'target_train_class_fmi_score',
% 'target_train_class_silhouette_score',
% 'target_train_class_dbi_score',
% 'target_train_class_chi_score',
% 'target_train_bnm_score',
% 'target_train_neg_snd_score',
% 'target_train_im_score',
% 'target_train_entropy_score',
% 'target_train_acc_score']

\section{Correlations} \label{sec:correlations}
\begin{sidewaysfigure}[p]
  \centering
  \includegraphics[width=0.32\linewidth]{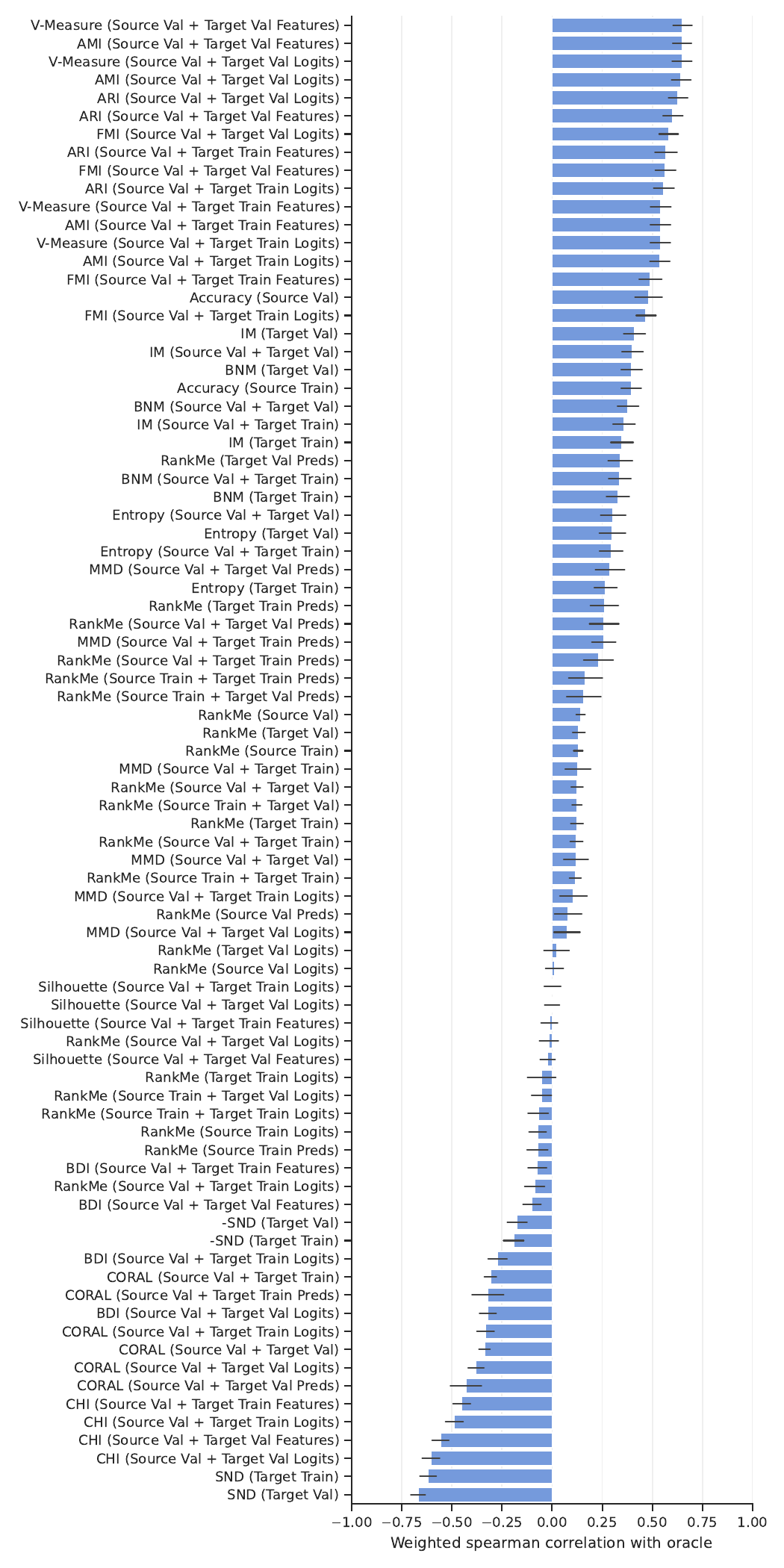}
  \includegraphics[width=0.32\linewidth]{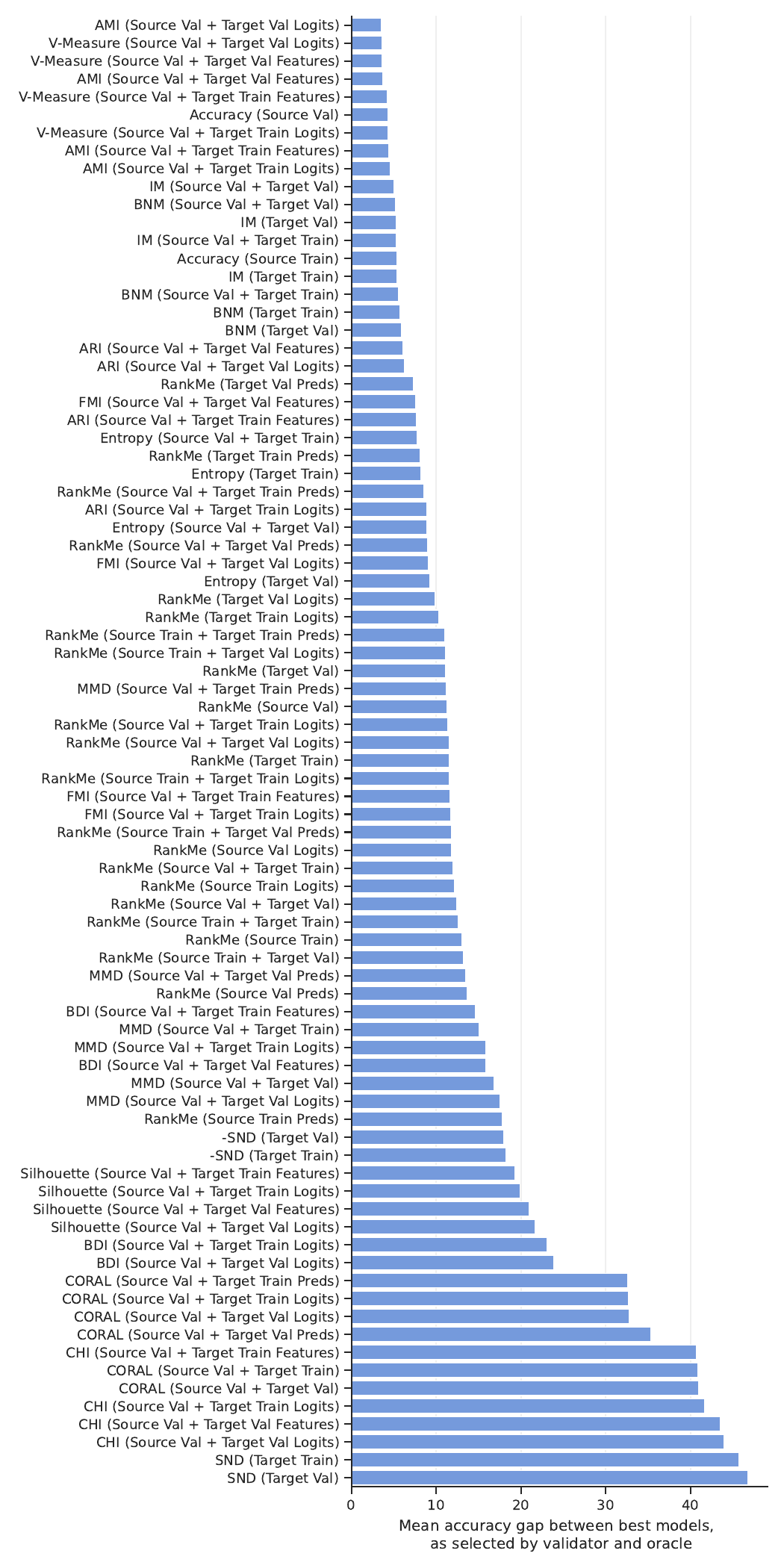}
  \includegraphics[width=0.32\linewidth]{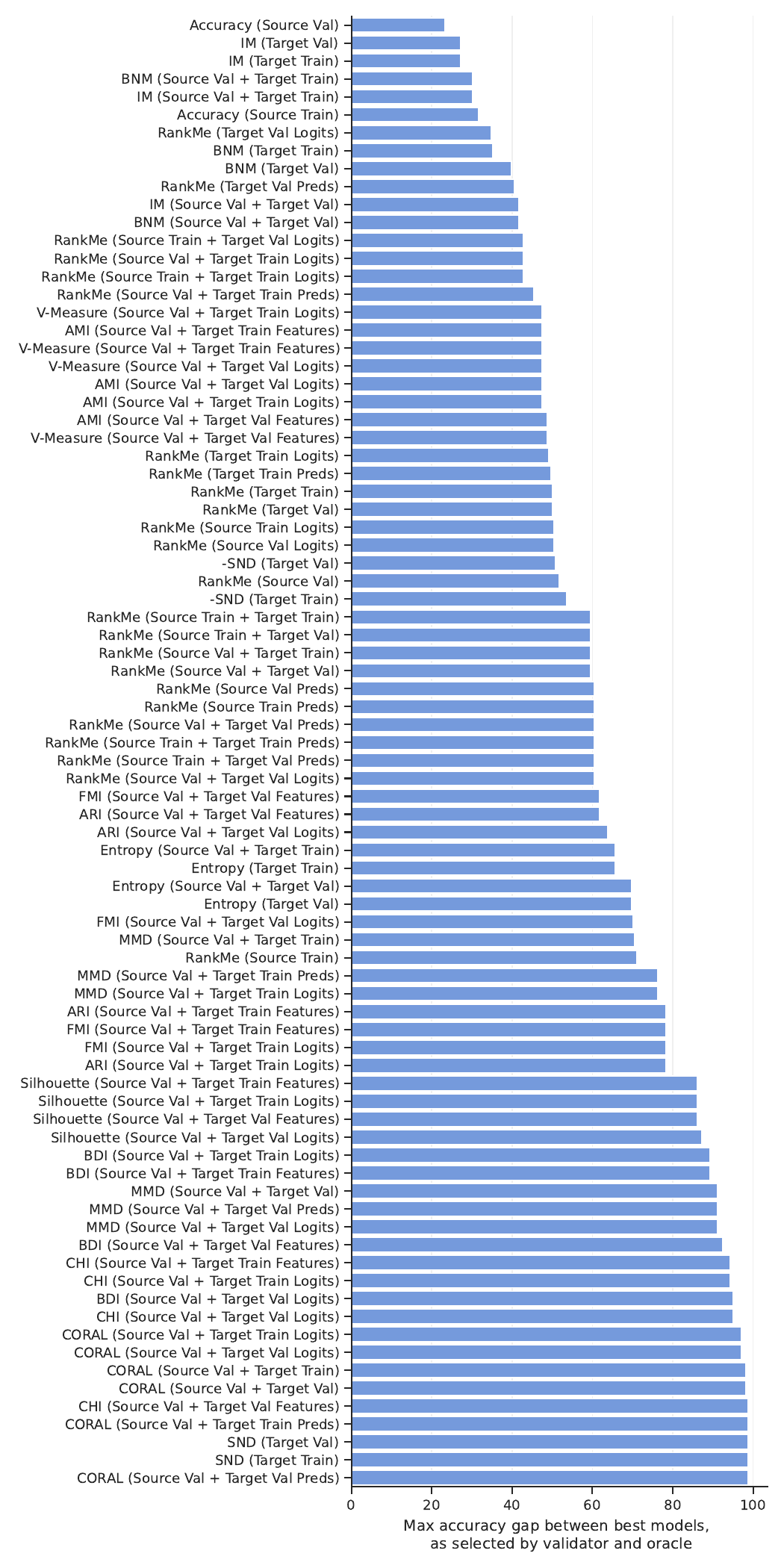}
  \caption{Left: Correlations with target test accuracy on UDA benchmarks. Error bars are standard error across domains. Middle: Average gap between the best model as selected by each validator and the oracle. Right: Maximum gap between the best model as selected by each validator and the oracle.}
  \label{fig:uda}
\end{sidewaysfigure}

\begin{sidewaysfigure}[p]
  \centering
  \includegraphics[width=0.32\linewidth]{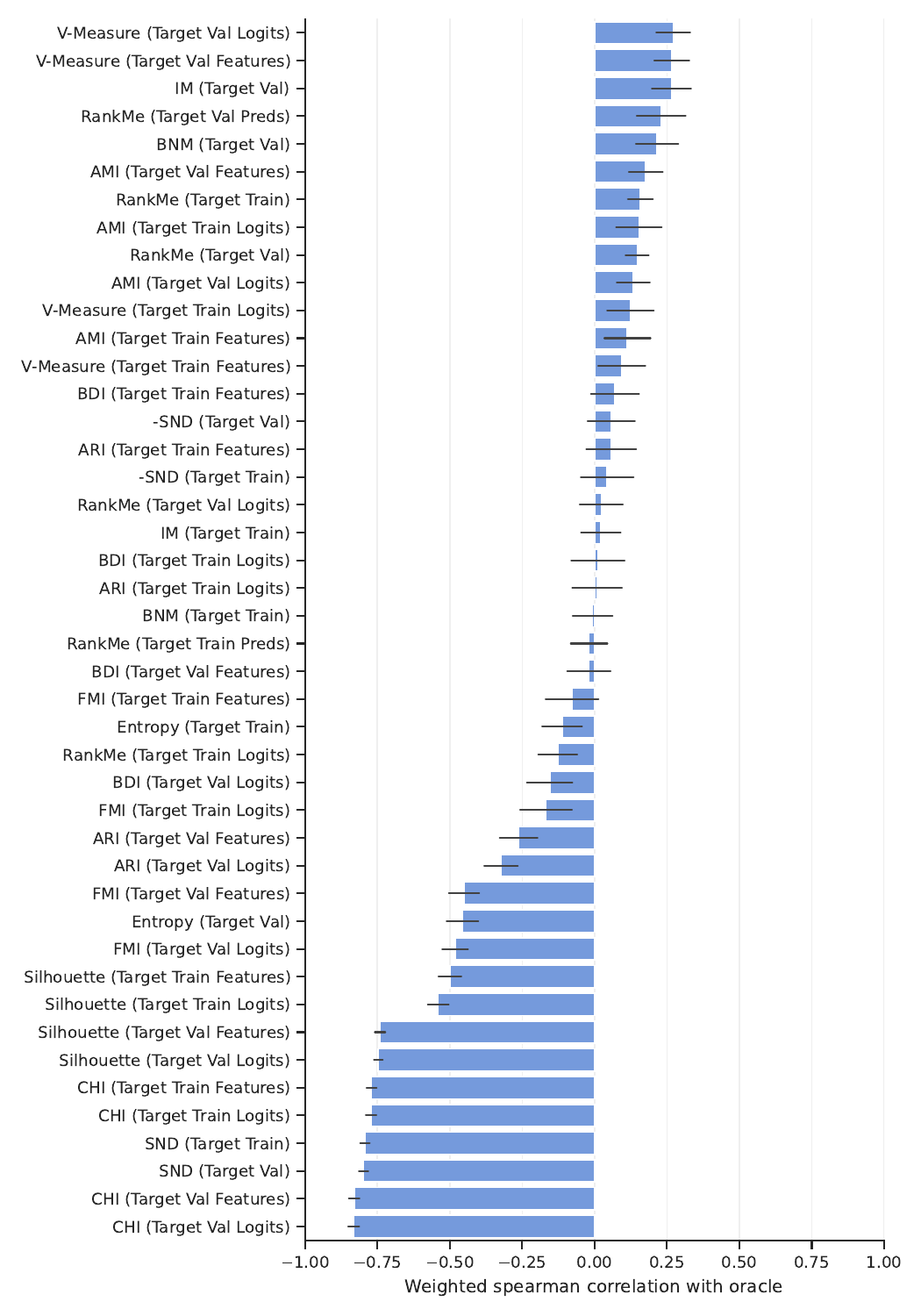}
  \includegraphics[width=0.32\linewidth]{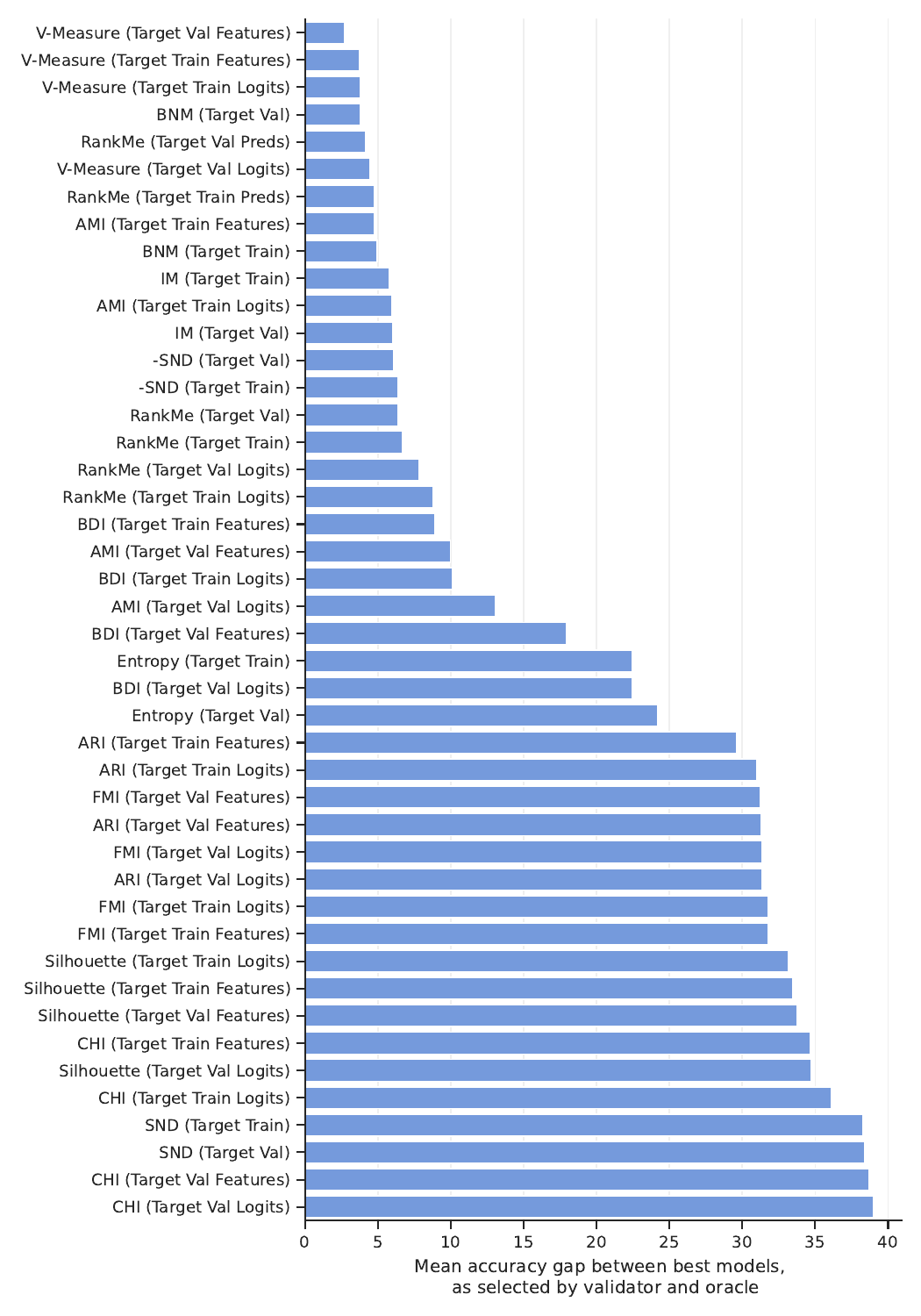}
  \includegraphics[width=0.32\linewidth]{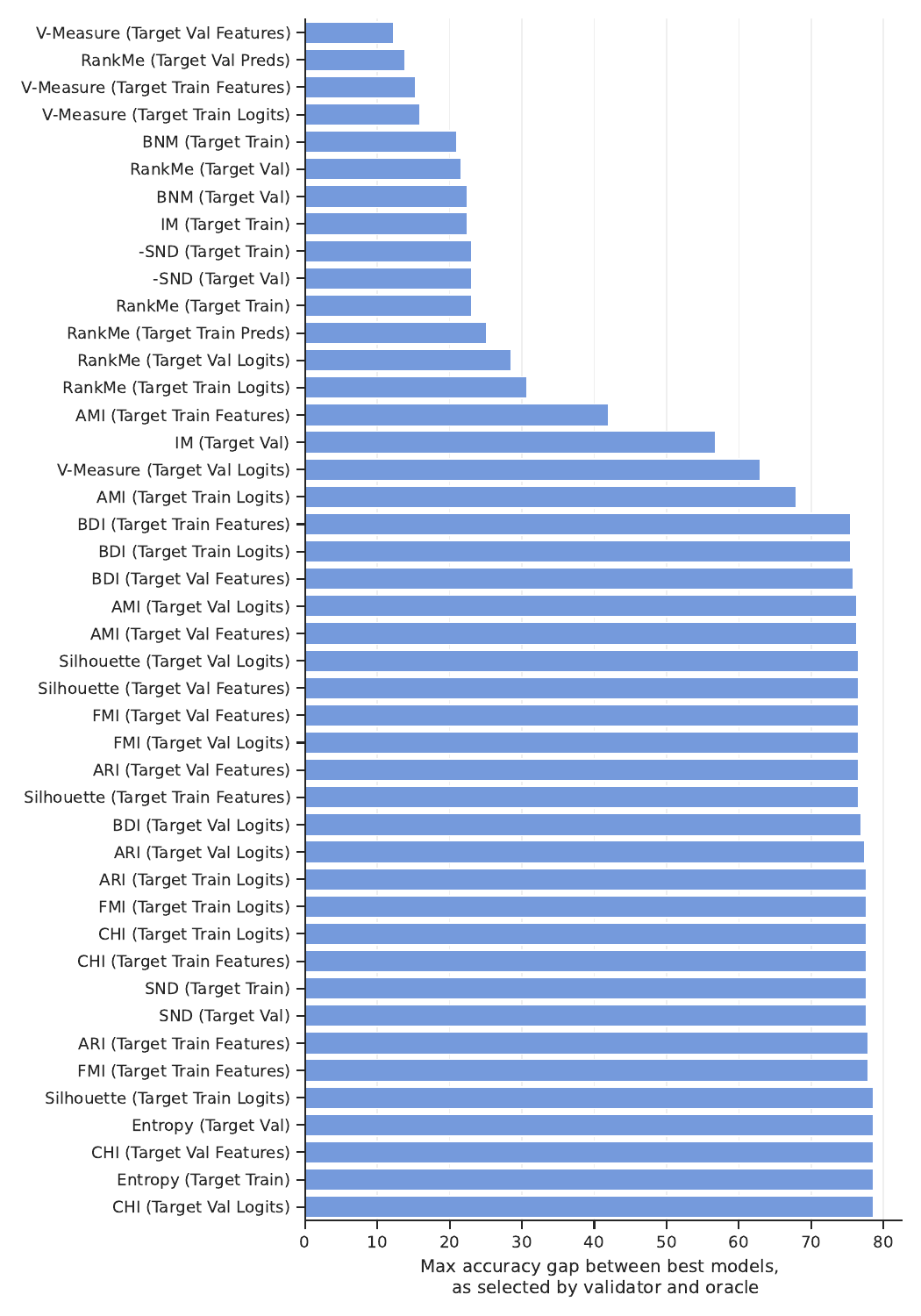}
  \caption{Left: Correlations with target test accuracy on SFDA benchmarks. Error bars are standard error across domains. Middle: Average gap between the best model as selected by each validator and the oracle. Right: Maximum gap between the best model as selected by each validator and the oracle.}
  \label{fig:sfda}
\end{sidewaysfigure}

\begin{sidewaysfigure}[p]
  \centering
  \includegraphics[width=0.32\linewidth]{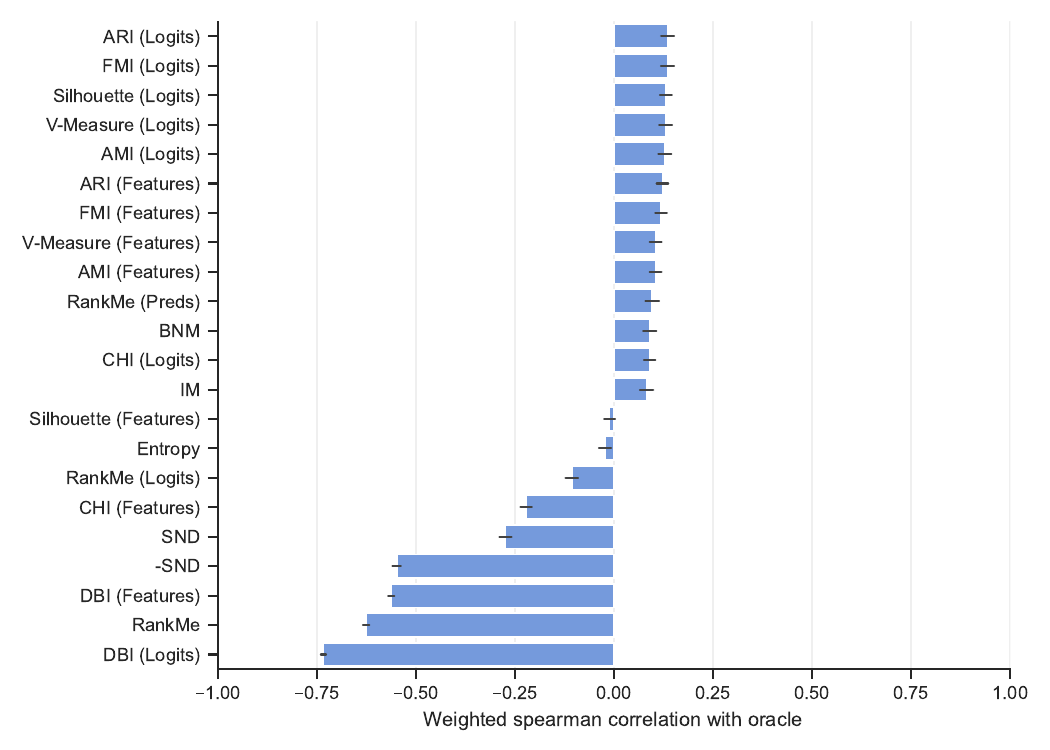}
  \includegraphics[width=0.32\linewidth]{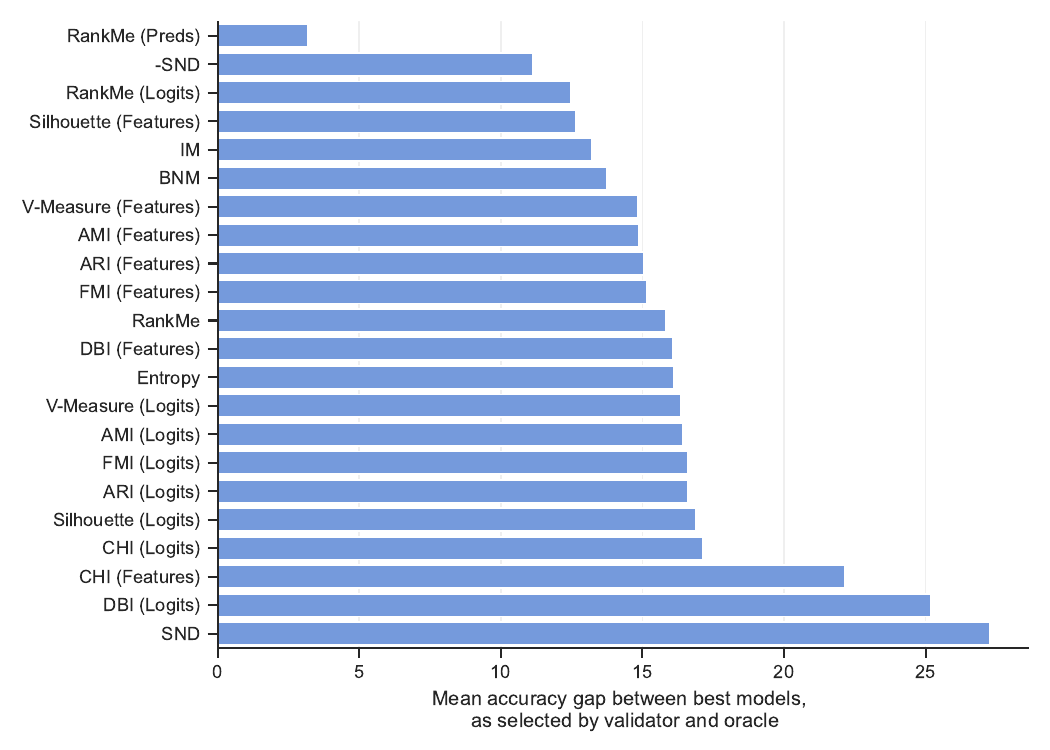}
  \includegraphics[width=0.32\linewidth]{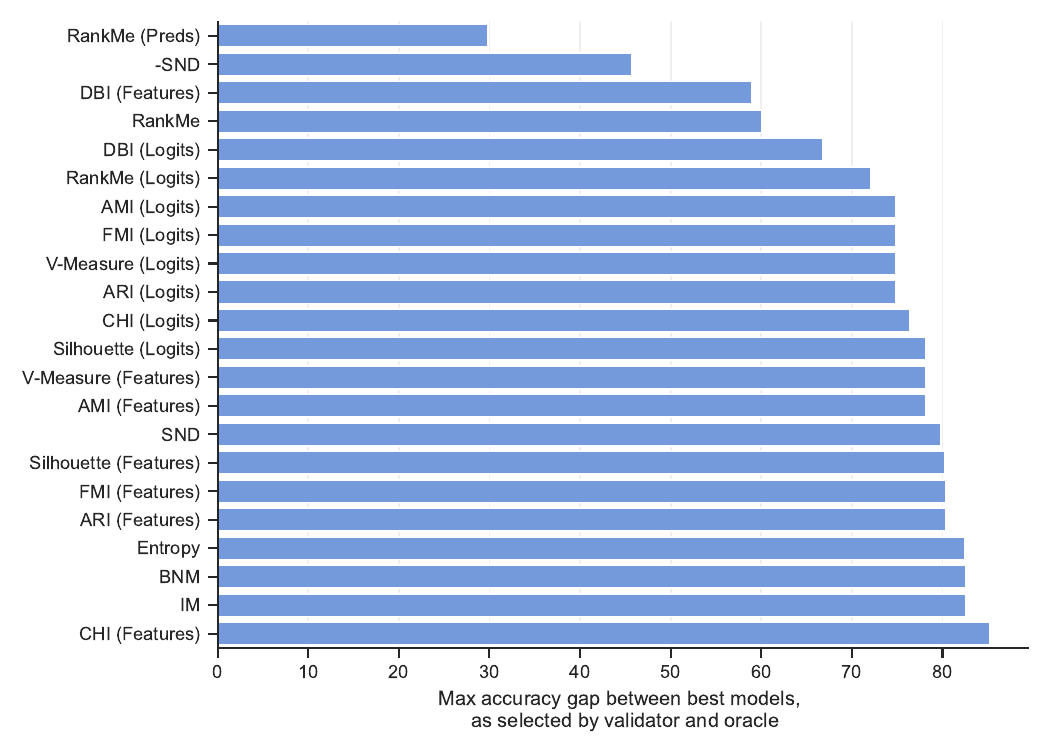}
  \caption{Left: Correlations with target test accuracy on TTA CIFAR10-C. Error bars are standard error across domains. Middle: Average gap between the best model as selected by each validator and the oracle. Right: Maximum gap between the best model as selected by each validator and the oracle.}
  \label{fig:tta_cifar10c}
\end{sidewaysfigure}

\begin{sidewaysfigure}[p]
  \centering
  \includegraphics[width=0.32\linewidth]{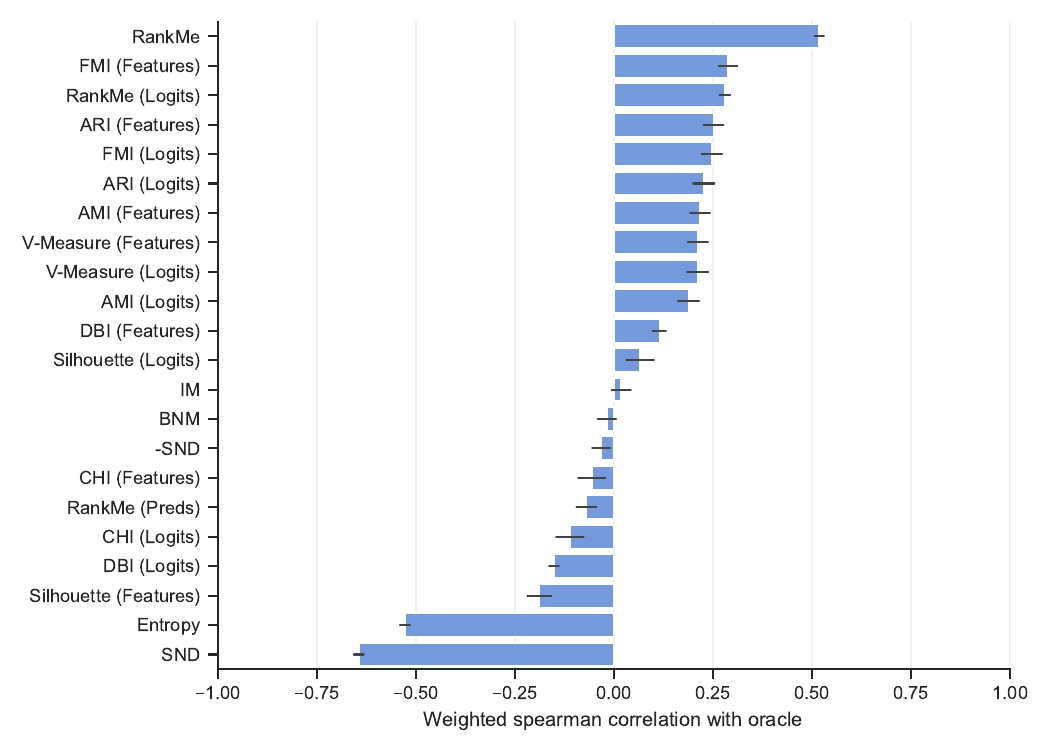}
  \includegraphics[width=0.32\linewidth]{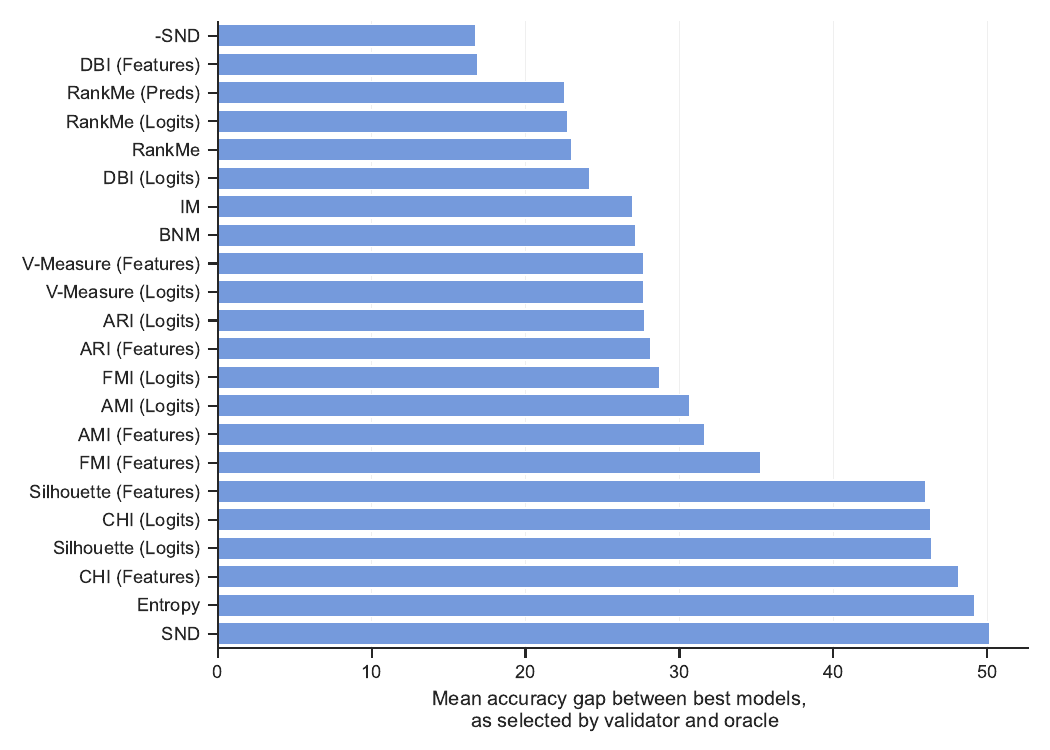}
  \includegraphics[width=0.32\linewidth]{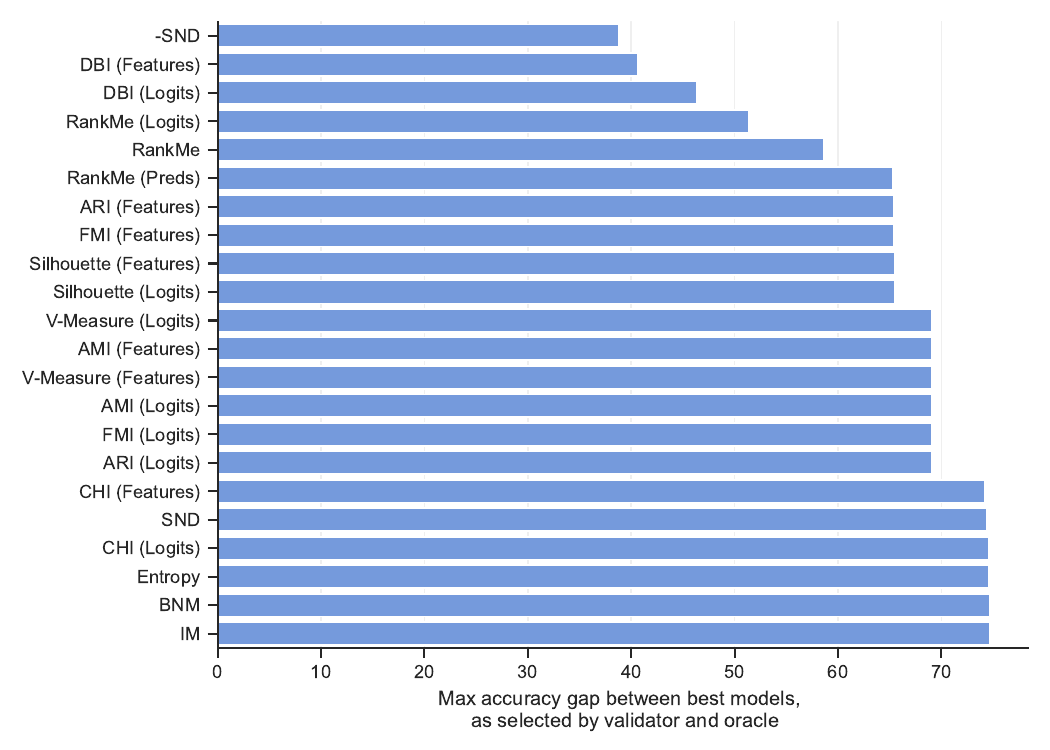}
  \caption{Left: Correlations with target test accuracy on TTA Office-Home. Error bars are standard error across domains. Middle: Average gap between the best model as selected by each validator and the oracle. Right: Maximum gap between the best model as selected by each validator and the oracle.}
  \label{fig:tta_officehome}
\end{sidewaysfigure}

\begin{figure}[p]
  \centering
  \includegraphics[width=0.7\linewidth]{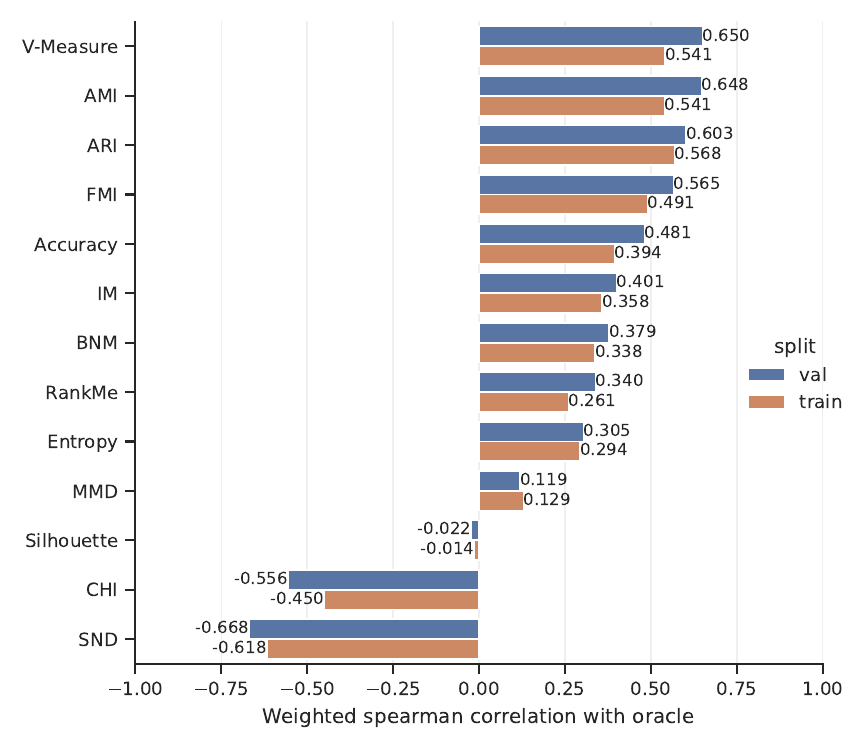}
  \caption{Comparison of split for evaluation of validation criteria in the UDA (classification) setting. We report the average weighted Spearman rank correlation between each validator and target test accuracy when using the following data splits for computing validators: (orange) target train data and (blue) target validation data.}
  \label{fig:trainOrValSplitCorr}
\end{figure}

Previous works have focused on identifying the validators that have the strongest correlation with the oracle \cite{musgrave2022benchmarkingUDA,Agostinelli2022HowSA}. Our main focus is on finding the ones that select the top-performing models and as we see in the main document, these different methods do not always lead to the same selection. For completeness, we include in \cref{fig:uda,fig:sfda,fig:tta_cifar10c,fig:tta_officehome} the weighted Spearman correlations (as used in \cite{musgrave2022benchmarkingUDA}) of all validators considered. Additionally, the comparison of train and val splits for validators in terms of correlation is shown in \cref{fig:trainOrValSplitCorr}.

\section{Compute Resources} \label{sec:compute}
The majority of experiments were run on an 8xA6000 internal cluster machine. The total number of algorithms we have trained and validated in this work is ~2,440. Assuming on average the training time is 1h per algorithm, this means 2,440 GPU hours have been used.

\end{document}